\documentclass[journal]{IEEEtai}

\usepackage[colorlinks,urlcolor=blue,linkcolor=blue,citecolor=blue]{hyperref}

\usepackage{color,array}

\usepackage{graphicx}
\usepackage{booktabs}
\usepackage{subcaption}
\usepackage{amssymb}
\usepackage{algorithm}
\usepackage{algorithmic}
\usepackage{amsmath}

\begin{document}

\title{Multi-Cue Anomaly Detection and Localization under Data Contamination}

\author{
\begin{minipage}[t]{0.45\textwidth}
\centering
\textbf{Anindya~Sundar~Das}\\
Department of Computing Science\\
Umeå University, Sweden\\
\texttt{aninsdas@cs.umu.se}
\end{minipage}
\hfill
\begin{minipage}[t]{0.45\textwidth}
\centering
\textbf{Monowar~Bhuyan}\\
Department of Computing Science\\
Umeå University, Sweden\\
\texttt{monowar@cs.umu.se}
\end{minipage}
}

\maketitle

\begin{abstract}
Visual anomaly detection in real-world industrial settings faces two major limitations. First, most existing methods are trained on purely normal data or on unlabeled datasets assumed to be predominantly normal, presuming the absence of contamination, an assumption that is rarely satisfied in practice. Second, they assume no access to labeled anomaly samples, limiting the model from learning discriminative characteristics of true anomalies. Therefore, these approaches often struggle to distinguish anomalies from normal instances, resulting in reduced detection and weak localization performance. In real-world applications, where training data are frequently contaminated with anomalies, such methods fail to deliver reliable performance. In this work, we propose a robust anomaly detection framework that integrates limited anomaly supervision into the adaptive deviation learning paradigm. We introduce a composite anomaly score that combines three complementary components: a deviation score capturing statistical irregularity, an entropy-based uncertainty score reflecting predictive inconsistency, and a segmentation-based score highlighting spatial abnormality. This unified scoring mechanism enables accurate detection and supports gradient-based localization, providing intuitive and explainable visual evidence of anomalous regions. Following the few-anomaly paradigm, we incorporate a small set of labeled anomalies during training while simultaneously mitigating the influence of contaminated samples through adaptive instance weighting. Extensive experiments on the MVTec and VisA benchmarks demonstrate that our framework outperforms state-of-the-art baselines and achieves strong detection and localization performance, interpretability, and robustness under various levels of data contamination. 
\end{abstract}




\section{Introduction}

Visual anomaly detection (VAD) plays an increasingly critical role in automated inspection tasks such as manufacturing quality control \cite{roth2022towards, liu2024deep}, medical image analysis \cite{le2023learning, wolleb2022diffusion}, and surveillance \cite{liu2018classifier}. 
The core challenge lies in characterizing normal patterns for a given domain and identifying deviations indicative of abnormality. 
Despite significant progress, many existing VAD methods rely on an unrealistic assumption: access to a large collection of clean and perfectly labeled normal samples. 
However, in real-world environments, training data often contain mislabeled instances, annotation errors, or unlabeled anomalies inherited from legacy systems, making data contamination unavoidable.

Although learning with noisy labels has been extensively studied in machine learning, through noise modeling \cite{hendrycks2018using}, soft-labeling strategies such as label smoothing \cite{szegedy2016rethinking}, and robust loss formulations \cite{ghosh2017robust, ma2020normalized}; these approaches are not inherently designed for anomaly detection.
Instance reweighting methods \cite{kumar2021constrained} offer a promising mechanism by down-weighting high-loss samples that are possibly corrupted \cite{arazo2019unsupervised, majidi2021exponentiated}, yet their systematic integration into VAD pipelines remains limited.
Moreover, beyond detection accuracy, many safety-critical applications increasingly require anomaly detectors to be interpretable, providing insight into both \emph{why} a sample is considered anomalous and \emph{where} abnormal evidence manifests spatially.

A further limitation in the VAD literature is the prevalent assumption that no labeled anomalies are available during training.
While large-scale anomaly annotation is indeed costly, many practical settings provide a small number of verified anomaly examples.
Deviation learning frameworks \cite{pang2019deep, pang2021explainable, das2023few} demonstrate that even a handful of labeled anomalies can substantially improve discriminative capability by explicitly modeling departures from normality.
However, existing deviation-based approaches typically focus on image-level anomaly detection and either do not explicitly account for training data contamination or do not provide robust, spatially grounded explanations; capabilities that are increasingly critical in industrial and medical deployments.

In this work, we advance deviation learning by introducing a \emph{contamination-robust and explainable anomaly detection framework} that extends prior approaches along three complementary dimensions.
\textbf{First}, we propose a \emph{multi-cue anomaly scoring strategy} that aggregates deviation magnitude, predictive uncertainty, and segmentation-derived spatial evidence, yielding more stable anomaly predictions under contaminated and ambiguous conditions.
\textbf{Second}, we adopt a \emph{few-shot anomaly supervision paradigm} that jointly leverages synthetic anomaly generation and a small set of real anomaly examples, while maintaining robustness to unknown contamination through adaptive instance reweighting.
\textbf{Third}, we derive \emph{gradient-based anomaly localization maps} that attribute anomaly relevance to spatial regions, providing interpretable explanations directly aligned with the learned anomaly scoring signals.

Extensive experiments on the MVTec AD and VisA benchmarks demonstrate that the proposed approach consistently outperforms competing methods across varying contamination levels, while simultaneously delivering reliable spatial localization and interpretable anomaly predictions suitable for real-world deployment.

\section{Related Work}
\label{sec:related_work}

AD approaches, introduced over the years \cite{cao2024survey, pang2021deep} are commonly categorized based on their learning paradigms and the level of supervision. A large class of approaches formulates anomaly detection as a one-class learning problem \cite{chen2022deep, ruff2020deep,tax2004support, ruff2018deep}, where models are trained using only normal samples and anomalies are identified as deviations from the learned distribution. Patch-level modeling and feature aggregation techniques further extend this idea by exploiting local representations \cite{yi2020patch}.

Reconstruction-based methods \cite{zavrtanik2021draem, hou2021divide, zavrtanik2021reconstruction} are optimized to accurately reconstruct normal images, using autoencoders \cite{sakurada2014anomaly, hinton2006reducing}, or generative adversarial networks \cite{ schlegl2019f, akcay2018ganomaly}. Anomalies are detected via elevated reconstruction error. Several other approaches \cite{zavrtanik2021reconstruction, ristea2022self} formulate anomaly detection as an inpainting task, in which image regions are randomly masked and neural networks are trained to recover the missing content.
Hybrid frameworks such as DR{\AE}M \cite{zavrtanik2021draem} combine generative synthesis with discriminative objectives to improve detection performance. 

Feature-based anomaly detection methods \cite{defard2021padim, rippel2021modeling, roth2022towards} leverage representations extracted from networks pre-trained on large-scale datasets such as ImageNet \cite{deng2009imagenet}. These approaches characterize normal feature distributions either through parametric models, such as Gaussian estimators, or through non-parametric similarity-based mechanisms, and compute anomaly scores based on distance or likelihood measures \cite{pang2021explainable}. Memory-based variants store normal feature prototypes and identify anomalies via nearest-neighbor similarity \cite{yi2020patch, roth2022towards}.

To address the limitations of purely unsupervised learning, several works incorporate limited labeled anomaly samples during training. Deviation learning methods \cite{pang2019deep, pang2021explainable, das2023few} introduce prior-driven objectives that enforce normal samples to follow a reference distribution while pushing anomalous samples toward statistically significant deviations. These approaches demonstrate improved discriminative power under limited anomaly supervision but are typically developed assuming clean training data and do not explicitly provide spatial localization or robustness to contamination.

Data augmentation-based strategies convert anomaly detection into a supervised learning problem by synthesizing pseudo-anomalies. Early augmentation techniques \cite{gidaris2018unsupervised, devries2017improved} are limited in modeling fine-grained defects, motivating more structured approaches such as CutPaste \cite{li2021cutpaste}, CutMix \cite{yun2019cutmix}, and Perlin noise-based anomaly synthesis \cite{zavrtanik2021draem, zavrtanik2022dsr, zhang2023destseg}. These methods improve detection performance but remain sensitive to contamination in the normal training set.

Learning under contaminated training data has also been explored, where unlabeled anomalies are present among normal samples \cite{wang2019effective, huyan2022unsupervised}. Blind training approaches treat contaminated data as clean and rely on robust statistics or sample prioritization \cite{wang2019effective}. Other methods iteratively identify and remove outliers during training \cite{yoon2021self}, while outlier exposure techniques explicitly leverage anomalous samples as auxiliary supervision \cite{hendrycks2018deep, qiu2022latent}. Existing methods, however, rarely combine limited anomaly supervision, contamination robustness, and spatially interpretable anomaly localization within a single framework.

\section{Problem Statement}
Let $\mathcal{X}_N=\{x_i\}_{i=1}^{N}$ denote the training set that is \emph{assumed normal} but may be contaminated, containing an unknown fraction $\epsilon$ of true anomalies. Following ADL~\cite{das2025adaptive}, we do not assume prior knowledge of $\epsilon$ and do not require access to a curated clean-only training set. In addition, we consider a very small set of $m$ labeled anomaly examples, denoted by $\mathcal{X}_A$.

For all samples $x_i \in \mathcal{X}_N$, we assign a nominal label $y_i = 0$, reflecting their assumed normality, while samples in $\mathcal{X}_A$ are labeled as anomalies with $y_i = 1$. We further synthetically generate a set of pseudo-anomalies $\mathcal{X}_{\hat{A}} = \{x_j\}_{j=1}^{M}$ to guide deviation learning, where each $x_j \in \mathcal{X}_{\hat{A}}$ is assigned the label $y_j = 1$.

Our goal is to learn (i) a robust anomaly score $s(x) \in \mathbb{R}$, derived from multiple complementary anomaly signals and remaining reliable under training data contamination, and (ii) a saliency-based localization map $\mathcal{H}(x) \in \mathbb{R}^{H \times W}$ that highlights anomalous regions and provides an interpretable explanation aligned with the predicted anomaly score.

\section{Proposed Approach}
\begin{figure*}[!ht]
\centering
  \includegraphics[width=0.9\linewidth]{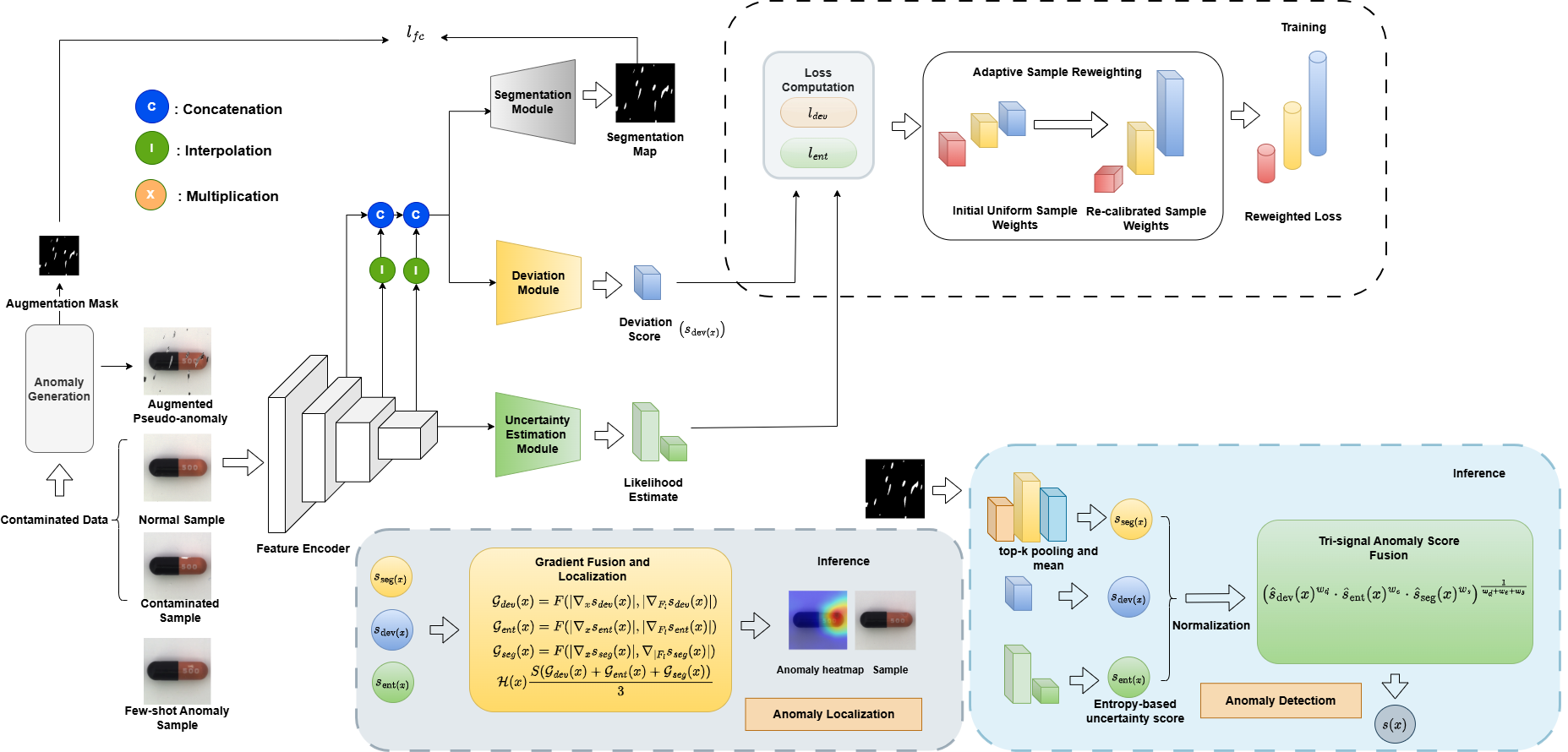}
  \caption{Overview of the proposed framework.
During training, a segmentation branch, a deviation module, and an uncertainty (entropy) module are learned using synthetic anomaly generation and limited anomaly supervision.
At inference, anomaly \emph{detection} is obtained via tri-signal score fusion, while anomaly \emph{localization} is produced by a separate gradient-based fusion module that aggregates signal-specific gradients to generate a spatial heatmap.
}
  \label{fig:architec}
\end{figure*}

This section presents our proposed framework for robust and explainable anomaly detection under limited anomaly supervision.
Building upon adaptive deviation learning, which produces a single deviation-based anomaly score optimized for robustness under contaminated training data, we incorporate additional complementary anomaly cues and explicit spatial explanation mechanisms. 
Our approach integrates multiple sources of anomaly evidence, jointly leverages synthetic anomaly generation and a small set of real anomaly samples, and produces both an image-level anomaly score and an interpretable spatial explanation, while preserving robustness to unknown data contamination during training. 
The overall structure of the proposed approach is illustrated in Fig.~\ref{fig:architec}. Our framework is organized around three tightly coupled components: \emph{contamination-robust deviation learning}, \emph{multi-cue anomaly scoring}, and \emph{gradient-based anomaly localization}.

\subsection{Contamination-Robust Deviation Learning Backbone}
\label{sec:RDL}
Our proposed framework builds upon the contamination-robust adaptive deviation learning paradigm introduced in \cite{das2025adaptive}, which we adopt as the training backbone of our method. This backbone enables reliable anomaly score learning in the presence of unknown data contamination by combining self-supervised pseudo-anomaly generation, deviation-based scoring, and adaptive sample reweighting. We briefly summarize the key components below for completeness, while retaining the original formulation without modification.

\paragraph{Synthetic Anomaly Generation}
Following \cite{zavrtanik2021draem, das2025adaptive}, we adopt a self-supervised pseudo-anomaly generation strategy to enable discriminative training under weak supervision. Given a normal image $I_n$, a synthetic anomaly mask $M_{\hat{a}}$ is generated using Perlin noise \cite{perlin1985image}, and an anomaly source image $I_{source}$ is sampled from an external dataset. The pseudo-anomalous image $I_{\hat{a}}$ is constructed as
\begin{equation}
I_{\hat{a}} = (1 - M_{\hat{a}}) \odot I_n + \beta (M_{\hat{a}} \odot I_{source}) + (1 - \beta)(M_{\hat{a}} \odot I_n)
\label{eq:pse_anomaly}
\end{equation}
where $\beta \in [0.1, 1.0]$ is a randomly sampled opacity coefficient and $\odot$ denotes element-wise multiplication. This procedure produces diverse, spatially irregular pseudo-anomalies that facilitate robust deviation learning.

\paragraph{Few-shot anomaly supervision}
In addition to the potentially contaminated training set, we incorporate a very small set of $m$ labeled anomaly examples as weak supervision. While synthetic pseudo-anomalies are effective for inducing deviation from normality, they may not faithfully capture the appearance or distribution of real anomalies in the target data, which can bias deviation learning toward artifacts of the generation process. A limited number of real anomaly samples therefore act as sparse but reliable anchors, guiding the deviation objective toward the true anomaly manifold. These few-shot anomalies do not aim to model the full anomaly distribution; rather, they reinforce the separation enforced by the deviation loss by consistently pushing genuine anomalies toward the upper tail of the score distribution, while preserving the open-set nature of anomaly detection.

\paragraph{Feature Encoder}
Input images are processed by a CNN-based feature encoder pretrained on ImageNet. Feature maps extracted from multiple intermediate layers are spatially aligned via interpolation and concatenated to form a unified representation
$\mathcal{F}_{\mathrm{Co}} \in \mathbb{R}^{C \times H \times W}$, where $H$ and $W$ denote the spatial height and width, and $C$ denotes the total number of channels across the selected layers. This multi-scale representation balances semantic abstraction and spatial fidelity, and serves as the shared input to all downstream learning modules.

\paragraph{Deviation-Based Anomaly Score Learning}
An anomaly scoring module maps the combined feature representation $\mathcal{F}_{\mathrm{Co}}(x_i)$ to patch-level anomaly scores
$\psi(x_i) = \{\psi(x_{ij})\}_{j=1}^{\tilde{N}}$. Following deviation learning, the image-level anomaly score is computed via top-$k$ pooling:
\begin{equation}
s_{\mathrm{dev}}(x_i) = \frac{1}{k} \sum_{x_{ij} \in \mathcal{P}_k(x_i)} \psi(x_{ij}), \quad k = \max\!\big(1,\lceil \rho\,\tilde{N}\rceil\big)
\end{equation}

where $\tilde{N}$ is the number of patches, $\rho\in(0,1]$ is the top-$k$ ratio, and
$\mathcal{P}_k(x_i)$ denotes the set of $k$ patches with the highest scores.
Deviation is measured relative to a reference distribution by computing the standardized score
$Z_{\mathrm{std}}(x_i) = (s_{\mathrm{dev}}(x_i) - \mu_{\mathcal{S}}) / \sigma_{\mathcal{S}}$,
where $\mu_{\mathcal{S}}$ and $\sigma_{\mathcal{S}}$ are estimated from samples drawn from a Gaussian prior.
To account for label noise induced by contamination, a soft-deviation loss is employed:
\begin{equation}
l_{\mathrm{soft}}(x_i) = (1 - p(x_i)) |Z_{\mathrm{std}}(x_i)| + p(x_i) \max(0, \gamma - Z_{\mathrm{std}}(x_i))
\label{eq:soft_dev}
\end{equation}
where $\gamma$ denotes a confidence margin, and $p(x_i)$ represents the estimated probability of $x_i$ being anomalous.

\paragraph{Auxiliary Classification and Segmentation}
An auxiliary classification branch, referred to as the \textit{Uncertainty Estimation Module} (Fig.~\ref{fig:architec}), predicts the anomaly probability $p(x_i)$ from high-level feature representations and provides soft supervision for deviation learning by estimating the likelihood that a given sample is anomalous.
The module is trained using a binary cross-entropy loss defined as
\begin{multline}
    l_{bce}(x_i)= -(1-y_i)\log(1-p(x_i)) -y_i\log(p(x_i)), \\
    \text{where} \quad y_i \in \{0, 1\}
    \label{eq:bce}
\end{multline}
In parallel, a segmentation module predicts pixel-level anomaly masks from $\mathcal{F}_{\mathrm{Co}}(x_i)$ and is trained using a focal loss $l_{fc}$ \cite{lin2017focal} against synthetic anomaly masks.

\paragraph{Adaptive Sample Importance Learning}
To mitigate the effect of unknown data contamination, we adopt the adaptive sample importance learning strategy from ADL, which dynamically reweights training samples based on their loss values. The key idea is that contaminated or mislabeled samples tend to incur higher losses and should therefore contribute less to the optimization.

Formally, given a per-sample loss $l(x_i)$, instance weights $w_i$ are obtained by solving the following constrained optimization problem at the minibatch level:
\begin{equation}
\min_{w \ge 0,\ \sum_i w_i = 1} \sum_i w_i l(x_i),
\quad \text{s.t. } \mathrm{Div}(w, u) \le \delta,
\label{eq:optimize}
\end{equation}
where $u$ denotes the uniform distribution and $\mathrm{Div}(\cdot,\cdot)$ is a divergence measure that controls how far the learned weights deviate from uniform weighting.

Following \cite{kumar2021constrained, das2025adaptive}, we employ the $\alpha$-divergence family as a unifying formulation, which subsumes commonly used divergences such as reverse-KL ($\alpha=0$) and KL ($\alpha=1$) to compute instance importance weights. For $\alpha \neq 1$, the resulting closed-form update is given by
\begin{equation}
\label{eq:weight_update}
w_i
=
\frac{
\left[(1-\alpha)l(x_i) + \lambda\right]_+^{\frac{1}{\alpha-1}}
}{
\sum_j
\left[(1-\alpha)l(x_j) + \lambda\right]_+^{\frac{1}{\alpha-1}}
},
\end{equation}
where $[\,\cdot\,]_+ = \max(\cdot,0)$ and $\lambda$ is a normalization constant.
The special case $\alpha \rightarrow 1$ recovers the KL-divergence update. The reweighting procedure is applied independently to the soft-deviation loss and the auxiliary classification loss.

By reweighting the learning objectives using the resulting importance weights, the overall training objective suppresses the influence of high-loss, potentially contaminated samples, while preserving discriminative learning from reliable data.

\subsection{Multi-cue Anomaly Scoring}
\label{sec:multicue}

Although the deviation backbone yields a strong anomaly signal, relying on a single score can be brittle when (i) the training set is severely contaminated, (ii) synthetic pseudo-anomalies only partially match real defect characteristics, or (iii) the decision boundary is ambiguous for visually subtle anomalies. To improve robustness at inference time without introducing additional training complexity, we propose a \emph{multi-cue} scoring strategy that aggregates three complementary sources of evidence: deviation magnitude, predictive uncertainty, and spatial anomaly activation.

\paragraph{Tri-signal extraction}
Given a test image $x$, a single forward pass of the trained network produces (i) a deviation-based anomaly score $s_{\mathrm{dev}}(x)$, (ii) an anomaly probability $p(x)\in(0,1)$ from the uncertainty estimation module, and (iii) a pixel-level anomaly mask $A(x)=\{a_j(x)\}_{j=1}^{HW}\in[0,1]^{H\times W}$ produced by the segmentation branch (Sec. \ref{sec:RDL}). We convert the probability output into an uncertainty score as follows:
\begin{align}
E(x) &= -p(x)\log p(x) - (1-p(x))\log(1-p(x)) \notag \\
s_{\mathrm{ent}}(x) &= \log\!\big(1 + E(x)\big)
\label{eq:ent_score}
\end{align}
which assigns higher values to uncertain predictions and lower values to confident predictions. To incorporate spatial evidence, we summarize the segmentation map by averaging the top-$k$ pixel scores:

\begin{equation}
s_{\mathrm{seg}}(x)
=
\frac{1}{k}
\sum_{u \in \mathcal{T}_k(x)} a_u(x),
\quad
k=\max\!\big(1,\lceil \rho\, H W\rceil\big)
\label{eq:seg_score}
\end{equation}

where $\rho\in(0,1]$ is a fixed top-$k$ ratio and
$\mathcal{T}_k(x)$ denotes the set of $k$ pixels with the highest values in $A(x)$.
The deviation, uncertainty, and segmentation scores operate on different scales and exhibit distinct tail behaviors. We therefore apply a deterministic, monotone normalization to map each cue into $[0,1]$ while
preserving their relative ordering.
\begin{equation}
\hat{s}_c(x)=\phi_c\!\big(s_c(x)\big)\in[0,1],
\qquad c\in\{\mathrm{dev},\mathrm{ent},\mathrm{seg}\}.
\label{eq:score_norm}
\end{equation}


\paragraph{Tri-signal Anomaly Score Fusion}
Lastly, we fuse the normalized scores ($\hat{s}_{\mathrm{dev}}(x)$, $\hat{s}_{\mathrm{ent}}(x)$, $\hat{s}_{\mathrm{seg}}(x)$) using a weighted geometric mean:
\begin{equation}
s(x)
=
\left(
\hat{s}_{\mathrm{dev}}(x)^{w_\mathrm{d}}
\hat{s}_{\mathrm{ent}}(x)^{w_\mathrm{e}}
\hat{s}_{\mathrm{seg}}(x)^{w_\mathrm{s}}
\right)^{\frac{1}{w_\mathrm{d}+w_\mathrm{e}+w_\mathrm{s}}}
\label{eq:fused_anomaly_score}
\end{equation}

where $w_{\mathrm{d}}, w_{\mathrm{e}}, w_{\mathrm{s}}\ge 0$ control the contribution of each score. This multi-cue fusion penalizes inconsistent evidence, mitigates spurious responses from individual cues and typically yields more stable anomaly detection under contamination.

\subsection{Gradient-Based Anomaly Localization}
\label{sec:localization}

Although the proposed multi-cue scoring strategy provides a robust image-level anomaly prediction, it does not directly indicate the spatial origin of anomalous patterns. 
To provide interpretable explanations, we employ a back-propagation-based anomaly localization mechanism that attributes anomaly relevance to spatial regions based on gradient sensitivity. 
This design builds upon gradient-based explanations used in deviation learning, and extends them to a multi-cue and contamination-robust setting by jointly attributing multiple anomaly signals.

\paragraph{Signal-specific gradient attribution}
Given a test image $x$, the trained network produces three anomaly-related scalar signals: the deviation-based score $s_{\mathrm{dev}}(x)$, the uncertainty-based score $s_{\mathrm{ent}}(x)$, and the segmentation-derived score $s_{\mathrm{seg}}(x)$. 
For each signal $c \in \{\mathrm{dev}, \mathrm{ent}, \mathrm{seg}\}$, we compute gradient-based attributions with respect to both the input image and an intermediate feature representation. 
\begin{equation}
g_c^{x}(x) = \left| \nabla_{x} \, s_c(x) \right|,
\qquad
g_c^{F}(x) = \left| \nabla_{F_l} \, s_c(x) \right|.
\label{eq:grad_signals}
\end{equation}

Here $F_l(x)\in\mathbb{R}^{C_l\times H_l\times W_l}$ denotes the activation tensor at layer $l$ (in our implementation, the last encoder block).

\paragraph{Cue-wise attribution map construction}
The raw gradients may differ in scale and spatial resolution. 
We therefore construct a cue-specific localization map by aggregating and normalizing the input-level and feature-level gradients:
\begin{equation}
\mathcal{G}_c(x) = \mathcal{F}\!\left( g_c^{x}(x),\, g_c^{F}(x) \right),
\label{eq:cue_map}
\end{equation}
where $\mathcal{F}(\cdot)$ denotes a generic processing operator that aggregates channel-wise gradient magnitudes, upsamples feature-level gradients to the input resolution, and applies per-map normalization.

\paragraph{Gradient fusion and smoothing}
To obtain a stable and coherent anomaly localization, the cue-specific maps are fused by equal averaging and spatial smoothing:
\begin{equation}
\mathcal{H}(x)
=
\mathcal{S}\!\left(
\frac{
\mathcal{G}_{\mathrm{dev}}(x)
+
\mathcal{G}_{\mathrm{ent}}(x)
+
\mathcal{G}_{\mathrm{seg}}(x)
}{3}
\right),
\label{eq:grad_fusion}
\end{equation}
where $\mathcal{S}(\cdot)$ denotes a spatial smoothing operator (e.g., Gaussian filtering) applied to suppress noise and produce a smooth anomaly localization map.

\subsection{Interpretability of Anomaly Predictions}
The proposed framework provides interpretability at multiple levels of the anomaly detection process.
At the score level, the deviation score measures how strongly a sample departs from the normal prior; predictive uncertainty captures inconsistencies induced by contaminated training data, remaining low for clean normal samples and elevated for anomalous ones; and the segmentation cue summarizes spatial evidence of abnormal regions.
At the localization level, gradient-based attribution reveals the sensitivity of individual anomaly cues to localized input perturbations, indicating which regions contribute most to elevated anomaly scores.
Finally, the fused localization map visually highlights the spatial origin of anomalous behavior.
Together, these components enable interpretable anomaly detection by explaining not only \emph{where} anomalies occur, but also \emph{why} the model assigns a high anomaly score. Unlike post-hoc standalone visualization techniques, these explanations are directly derived from the learned anomaly scoring signals and closely aligned with the model’s decision process.

\begin{algorithm}[!ht]
\caption{Multi-Cue Anomaly Scoring with Limited Anomaly Supervision under Data Contamination}
\label{alg:proposed_adl_ext}
\begin{flushleft}
\textbf{Input:}
$\mathcal{X}_N$: contaminated nominal training set (unknown contamination ratio $\epsilon$);\;
$\mathcal{X}_{\hat{A}}$: synthetically generated pseudo-anomalies;\;
$\mathcal{X}_A$: few-shot real anomalies.\;
Training set $\mathcal{X}=\mathcal{X}_N \cup \mathcal{X}_{\hat{A}} \cup \mathcal{X}_A$.

\textbf{Output:}
final anomaly score $s(x)$ and localization map $\mathcal{H}(x)$ for a test sample $x$.

\textbf{Hyperparameters:}
$\gamma$ (margin in soft-deviation, Eq. \ref{eq:soft_dev});\;
$\alpha$ and $\lambda$ (divergence reweighting, Eq. \ref{eq:weight_update});\;
burn-in $burn\_in$;\;
top-$k$ ratio $\rho$;\;
fusion weights $(w_\mathrm{d},w_\mathrm{e},w_\mathrm{s})$.
\end{flushleft}

\begin{algorithmic}[1]
\STATE Initialize model parameters $\Theta$.
\FOR{$e=1$ to $E$}
  \FOR{$b=1$ to $num\_of\_batches$}
    \STATE Sample minibatch $\{(x_i, y_i, M_i)\}_{i=1}^{B}$ from $\mathcal{X}$.
    \STATE Sample reference scores $\mathcal{S}=\{s_j\}_{j=1}^m$ with $s_j\sim \mathcal{N}(0,1)$; compute $\mu_{\mathcal{S}},\sigma_{\mathcal{S}}$.
    \STATE Forward pass: $(A(x_i), s_{\mathrm{dev}}(x_i), p(x_i)) \leftarrow f_{\Theta}(x_i)$.
    \STATE Compute $l_{\mathrm{soft}}(x_i)$, $l_{\mathrm{bce}}(x_i)$, $l_{fc}(x_i)$.
    \IF{$e > burn\_in$}
      \STATE Compute instance weights $w^{(1)}_i$ for $l_{\mathrm{soft}}(x_i)$ and $w^{(2)}_i$ for $l_{\mathrm{bce}}(x_i)$ using Eq. \ref{eq:weight_update}
    \ELSE
      \STATE Set uniform weights $w^{(1)}_i=w^{(2)}_i=\frac{1}{B}\;\; \forall i$.
    \ENDIF
    \STATE Compute reweighted training objective:
    \[
      \mathcal{L} = \sum_i w^{(1)}_i\,l_{\mathrm{soft}}(x_i)\;+\;\sum_i w^{(2)}_i\,l_{\mathrm{bce}}(x_i)\;+\;\sum_i l_{fc}(x_i).
    \]
    \STATE Update $\Theta$ using $\nabla_{\Theta}\mathcal{L}$ (Adam).
  \ENDFOR
\ENDFOR

\STATE \textbf{Inference for a test sample $x$:}
\STATE Forward pass: $(A(x), s_{\mathrm{dev}}(x), p(x)) \leftarrow f_{\Theta}(x)$.
\STATE Compute uncertainty score $s_{\mathrm{ent}}(x)$ (Eq. \ref{eq:ent_score}), segmentation score $s_{\mathrm{seg}}(x)$ (Eq. \ref{eq:seg_score}). 
\STATE Fuse signals by weighted geometric mean to get anomaly score $s(x)$ using Eq. \ref{eq:fused_anomaly_score}
\STATE Obtain localization $\mathcal{H}(x)$ via multi-cue gradients as shown in Eq.~\ref{eq:grad_signals}, \ref{eq:cue_map}, \ref{eq:grad_fusion}.

\end{algorithmic}
\end{algorithm}

\section{Evaluation}

\subsection{Experimental Setup}

\subsubsection{Datasets}
We conduct experiments on two widely used benchmark datasets for VAD: \emph{MVTec AD} \cite{bergmann2019mvtec} and \emph{VisA} \cite{zou2022spot}. 
MVTec AD contains 15 industrial categories spanning both texture and object-based industrial inspection tasks, totaling 5,354 images, of which  1725 images are reserved for testing. 
For each category, the training split contains only nominal samples, while the test split includes both normal and anomalous instances with pixel-level ground-truth masks.
VisA consists of 12 categories, providing 8,659 normal training images and a test set composed of 962 normal and 1,200 anomalous samples, each annotated with pixel-level anomaly masks.

\subsubsection{Contaminated Training Protocol}
To simulate real-world deployment scenarios, we deliberately introduce contamination into the training data.
Specifically, a fraction $\epsilon$ of the normal training set is contaminated with anomalous samples that are treated as normal during training.
Since both datasets provide clean training splits, we follow prior work \cite{das2025adaptive, qiu2022latent} and construct contaminated samples by drawing anomalous instances from the test set and perturbing them using additive zero-mean Gaussian noise with high variance.
This strategy injects anomalies directly into the image space, enabling controlled evaluation under varying contamination ratios.
All competing methods are trained using the same contaminated data protocol to ensure fair comparison, except for LOE \cite{qiu2022latent}, which follows its original latent-space contamination mechanism.

\subsubsection{Implementation Details}
The proposed framework is trained following the architecture shown in Fig.~\ref{fig:architec}.
Input images are resized and center-cropped to a resolution of $256\times256$.
We employ an ImageNet-pretrained ResNet-18 backbone \cite{he2016deep} and extract feature maps from intermediate convolutional layers (Layers 2–4), which are spatially aligned and concatenated to form a shared representation.
This representation is jointly consumed by the deviation scoring head, uncertainty estimation module, and segmentation branch.

Synthetic anomalies are generated using external texture images from the Describable Textures Dataset (DTD) \cite{cimpoi2014describing}.
The network is optimized using the Adam optimizer with a learning rate of $2\times10^{-4}$ and trained for 25 epochs on both datasets.
Image-level deviation scores are computed via top-$k$ pooling with ratio $\rho=0.1$.
For the segmentation branch, we compute an image-level segmentation cue by averaging
the top-$k$ highest values in the predicted anomaly map, corresponding to $10\%$ of the
spatial locations.
Reference scores are sampled from a standard normal distribution with 5,000 samples, and the confidence margin in the deviation loss is set to $\gamma=5$.
Adaptive instance reweighting is implemented using the $\alpha$-divergence formulation with $\alpha=0.1$ and $\lambda=0.1$, selected via grid search. For the tri-signal anomaly score fusion, the weighting coefficients for the deviation, entropy, and segmentation cues are set to $w_\mathrm{d} = 0.55$, $w_\mathrm{e} = 0.10$, and $w_\mathrm{s} = 0.35$, respectively, also determined through grid search.

In addition to synthetic pseudo-anomalies, we incorporate a small number of real anomaly samples as weak supervision.
To emulate practical deployment scenarios, a fixed number of anomalous instances are randomly sampled from the test split and included in the training set with anomaly labels.
Unless stated otherwise, we fix the number of few-shot true anomalies to $m=10$ per category, reserve all remaining anomalous samples exclusively for evaluation.

\subsubsection{Baselines and Evaluation Metrics}
We compare the proposed method against several state-of-the-art anomaly detection approaches, including PatchCore \cite{roth2022towards}, PaDiM \cite{defard2021padim},  DestSeg \cite{zhang2023destseg}, DR{\AE}M \cite{zavrtanik2021draem}, LOE \cite{qiu2022latent}, ADL \cite{das2025adaptive}.
Official implementations are used whenever available, and all baselines are adapted to operate under contaminated training data to ensure consistency across methods.
For LOE, we employ the authors’ released \emph{soft}-LOE implementation, which uses a ResNet-152 backbone. Since the original ADL framework does not explicitly support pixel-level anomaly maps, we derive localization results by applying gradient-based saliency with respect to the deviation-based anomaly score.

Detection performance is evaluated using image-level metrics, including the Area Under the Receiver Operating Characteristic curve (AUROC) and the Area Under the Precision-Recall curve (AUPRC).
AUROC captures the trade-off between true-positive and false-positive rates, while AUPRC is particularly informative for anomaly detection due to the severe class imbalance between normal and anomalous samples.
Higher values indicate better detection performance for both metrics. To assess anomaly localization quality, we additionally report pixel-level AUROC computed between the predicted anomaly maps and ground-truth segmentation masks.
This metric evaluates the model’s ability to accurately highlight anomalous regions at the spatial level, complementing image-level detection performance.

\subsection{Results and Analysis}

In this section, we compare the proposed method with state-of-the-art anomaly detection and localization baselines under varying contamination ratios on MVTec AD and VisA. We report image-level detection AUROC and pixel-level localization AUROC to characterize performance across datasets and contamination levels.

\begin{table*}[ht]
\centering
\caption{Quantitative comparison of image-level AUROC for anomaly detection on MVTec AD and VisA under progressively contaminated training data. The best and second-best results are indicated by \textbf{bold} and \underline{underline}, respectively.}
\scalebox{0.8}{
\begin{tabular}{llllllll}
\toprule
Model & PatchCore & PaDiM & DestSeg & DR{\AE}M & LOE & ADL & Ours \\
\midrule
MVTec Classes & 10\%/15\%/20\% & 10\%/15\%/20\% & 10\%/15\%/20\% &
10\%/15\%/20\% & 10\%/15\%/20\% & 10\%/15\%/20\% & 10\%/15\%/20\% \\
\midrule
carpet     & 0.781/0.681/0.662 & 0.856/0.817/0.806 & 0.941/0.828/0.758 & 0.801/0.758/0.731 & 0.938/0.925/\underline{0.922} & \underline{0.942}/\underline{0.931}/0.887 & \textbf{0.996}/\textbf{0.991}/\textbf{0.989} \\
bottle     & 0.889/0.831/0.757 & 0.828/0.738/0.661 & \textbf{1.000}/\textbf{1.000}/\textbf{1.000} & 0.965/0.958/0.949 & 0.982/0.981/0.979 & \textbf{1.000}/\textbf{1.000}/\underline{0.999} & \underline{0.999}/\underline{0.999}/0.998 \\
cable      & \underline{0.880}/0.797/0.737 & 0.651/0.495/0.344 & 0.715/0.561/0.418 & 0.685/0.637/0.497 & 0.845/\underline{0.829}/\underline{0.813} & 0.848/0.767/0.737 & \textbf{0.934}/\textbf{0.923}/\textbf{0.853} \\
capsule    & 0.859/0.782/0.781 & 0.813/0.758/0.703 & \underline{0.938}/\textbf{0.937}/\textbf{0.937} & \textbf{0.944}/0.891/\underline{0.889} & 0.789/0.782/0.765 & 0.864/0.859/0.825 & 0.932/\underline{0.930}/0.885 \\
zipper     & 0.873/0.791/0.789 & 0.768/0.701/0.643 & \underline{0.998}/\underline{0.998}/\underline{0.998} & \textbf{0.999}/0.990/0.990 & 0.881/0.875/0.871 & 0.958/0.953/0.952 & \textbf{0.999}/\textbf{0.999}/\textbf{0.999} \\
wood       & 0.745/0.642/0.567 & 0.685/0.550/0.485 & 0.921/0.917/0.895 & 0.714/0.645/0.641 & 0.937/0.913/0.895 & \textbf{0.999}/\textbf{0.998}/\textbf{0.997} & \underline{0.991}/\underline{0.989}/\underline{0.950} \\
transistor & 0.839/0.827/0.701 & 0.408/0.215/0.196 & 0.809/0.678/0.581 & 0.435/0.297/0.294 & \underline{0.841}/\underline{0.839}/\underline{0.835} & 0.825/0.805/0.721 & \textbf{0.922}/\textbf{0.875}/\textbf{0.874} \\
toothbrush & 0.858/0.742/0.732 & 0.775/0.656/0.489 & 0.885/0.882/0.811 & \textbf{0.911}/\textbf{0.902}/0.789 & \underline{0.909}/\underline{0.901}/\textbf{0.889} & 0.855/0.824/0.767 & 0.904/0.863/\underline{0.838} \\
tile       & 0.838/0.755/0.741 & 0.696/0.533/0.367 & \underline{0.962}/0.941/0.871 & 0.942/0.929/0.805 & \textbf{0.992}/\textbf{0.991}/\textbf{0.991} & 0.953/0.928/0.921 & 0.948/\underline{0.945}/\underline{0.942} \\
screw      & 0.815/0.739/0.678 & 0.601/0.445/0.293 & 0.855/\textbf{0.852}/\textbf{0.847} & \textbf{0.894}/0.805/0.801 & 0.551/0.545/0.532 & 0.844/\underline{0.841}/0.777 & \underline{0.857}/0.837/\underline{0.825} \\
pill       & 0.809/0.739/0.671 & 0.739/0.616/0.504 & \underline{0.925}/\underline{0.919}/\underline{0.915} & 0.895/0.874/0.872 & 0.752/0.738/0.714 & 0.917/0.886/0.869 & \textbf{0.963}/\textbf{0.958}/\textbf{0.917} \\
metal\_nut & 0.848/0.755/0.679 & 0.736/0.588/0.409 & 0.975/0.913/0.858 & 0.977/0.938/0.825 & 0.738/0.712/0.709 & \underline{0.982}/\underline{0.971}/\underline{0.865} & \textbf{0.985}/\textbf{0.982}/\textbf{0.977} \\
leather    & 0.781/0.721/0.685 & 0.950/0.941/0.935 & \textbf{1.000}/\textbf{1.000}/\textbf{1.000} & \textbf{1.000}/\textbf{1.000}/\textbf{1.000} & 0.988/0.984/0.972 & \underline{0.999}/\underline{0.999}/0.988 & \textbf{1.000}/0.995/\underline{0.994} \\
hazelnut   & 0.639/0.621/0.619 & 0.416/0.346/0.314 & \underline{0.994}/\underline{0.978}/\underline{0.976} & 0.981/0.961/0.875 & 0.959/0.951/0.941 & 0.985/\underline{0.978}/0.972 & \textbf{1.000}/\textbf{0.997}/\textbf{0.979} \\
grid       & 0.621/0.605/0.521 & 0.486/0.334/0.303 & \textbf{0.999}/\textbf{0.995}/\textbf{0.994} & \underline{0.979}/0.917/0.795 & 0.475/0.437/0.411 & 0.896/0.885/0.825 & 0.930/\underline{0.927}/\underline{0.917} \\
\textbf{Average} &
0.805/0.735/0.688 &
0.694/0.582/0.497 &
\underline{0.928}/0.893/0.857 &
0.875/0.833/0.784 &
0.838/0.827/0.816 &
0.924/\underline{0.908}/\underline{0.873} &
\textbf{0.957}/\textbf{0.947}/\textbf{0.929} \\
\midrule
VisA Classes & 10\%/15\%/20\% & 10\%/15\%/20\% & 10\%/15\%/20\% &
10\%/15\%/20\% & 10\%/15\%/20\% & 10\%/15\%/20\% & 10\%/15\%/20\% \\
\midrule
capsules   & 0.551/0.488/0.481 & 0.290/0.130/0.187 & 0.661/0.631/0.628 & 0.465/0.441/0.411 & 0.625/0.575/0.538 & \underline{0.752}/\textbf{0.743}/\textbf{0.723} & \textbf{0.785}/\underline{0.733}/\underline{0.701} \\
candle     & 0.749/0.741/0.735 & 0.856/0.786/0.725 & \underline{0.917}/\underline{0.909}/\underline{0.900} & 0.819/0.733/0.664 & 0.866/0.845/0.811 & 0.913/0.904/0.880 & \textbf{0.972}/\textbf{0.971}/\textbf{0.967} \\
cashew     & 0.772/0.755/0.725 & 0.856/0.791/0.662 & 0.850/0.839/0.825 & 0.433/0.311/0.250 & \underline{0.941}/\underline{0.929}/\underline{0.919} & 0.909/0.899/0.871 & \textbf{0.971}/\textbf{0.967}/\textbf{0.952} \\
chewinggum & 0.717/0.585/0.541 & 0.836/0.824/0.623 & 0.849/0.834/0.829 & 0.809/0.753/0.711 & \underline{0.961}/\underline{0.959}/\underline{0.951} & 0.928/0.921/0.913 & \textbf{0.977}/\textbf{0.964}/\textbf{0.962} \\
macaroni1  & 0.691/0.682/0.663 & 0.731/0.632/0.553 & 0.746/0.737/0.731 & \underline{0.861}/\underline{0.847}/\underline{0.825} & 0.671/0.658/0.599 & 0.808/0.739/0.728 & \textbf{0.951}/\textbf{0.948}/\textbf{0.945} \\
macaroni2  & 0.589/0.537/0.504 & 0.119/0.115/0.113 & 0.652/0.631/0.605 & \underline{0.724}/\underline{0.709}/\underline{0.681} & 0.513/0.505/0.431 & 0.609/0.596/0.554 & \textbf{0.737}/\textbf{0.726}/\textbf{0.722} \\
fryum      & 0.769/0.714/0.695 & 0.723/0.651/0.566 & 0.835/0.831/0.810 & 0.749/0.665/0.558 & 0.756/0.748/0.738 & \underline{0.934}/\underline{0.927}/\underline{0.913} & \textbf{0.979}/\textbf{0.952}/\textbf{0.950} \\
pipe\_fryum& 0.759/0.631/0.594 & 0.846/0.798/0.692 & 0.925/0.911/0.901 & \underline{0.957}/\underline{0.945}/\underline{0.942} & 0.813/0.809/0.803 & 0.917/0.911/0.877 & \textbf{0.973}/\textbf{0.967}/\textbf{0.961} \\
pcb1       & 0.735/0.729/0.721 & 0.832/0.728/0.649 & \underline{0.851}/\underline{0.841}/\textbf{0.829} & 0.211/0.144/0.119 & 0.817/0.795/0.778 & 0.821/0.816/0.783 & \textbf{0.901}/\textbf{0.882}/\underline{0.801} \\
pcb2       & 0.715/0.709/0.695 & 0.769/0.653/0.567 & 0.791/0.765/0.761 & 0.235/0.195/0.129 & \underline{0.849}/0.787/0.739 & 0.824/\underline{0.804}/\underline{0.791} & \textbf{0.889}/\textbf{0.886}/\textbf{0.840} \\
pcb3       & 0.665/0.651/0.644 & 0.746/0.633/0.537 & \textbf{0.922}/\textbf{0.895}/\textbf{0.868} & 0.707/0.604/0.455 & 0.798/0.759/0.713 & 0.814/\underline{0.810}/\underline{0.792} & \underline{0.820}/0.770/0.762 \\
pcb4       & 0.891/0.880/0.869 & 0.930/0.846/0.779 & \underline{0.964}/\textbf{0.961}/\textbf{0.945} & 0.805/0.757/0.778 & \textbf{0.967}/0.931/0.911 & 0.912/0.863/0.857 & 0.943/\underline{0.937}/\underline{0.926} \\
\textbf{Average} &
0.717/0.675/0.656 &
0.711/0.632/0.554 &
0.830/0.815/0.803 &
0.648/0.592/0.544 &
0.798/0.775/0.744 &
\underline{0.845}/\underline{0.828}/\underline{0.807} &
\textbf{0.908}/\textbf{0.892}/\textbf{0.874} \\
\bottomrule
\label{tab:image_roc}
\end{tabular}
}
\end{table*}


\begin{table*}[!t]
\centering
\caption{Quantitative comparison of pixel-level AUROC for anomaly localization on MVTec AD and VisA under progressively contaminated training data. The best and second-best results are indicated by \textbf{bold} and \underline{underline}, respectively.}
\label{tab:pixel_roc}
\scalebox{0.8}{
\begin{tabular}{l*{7}{c}}
\toprule
Model & PatchCore & PaDiM & DestSeg & DR{\AE}M & LOE & ADL & Ours \\
\midrule
MVTec Classes & 10\%/15\%/20\% & 10\%/15\%/20\% & 10\%/15\%/20\% & 10\%/15\%/20\% & 10\%/15\%/20\% & 10\%/15\%/20\% & 10\%/15\%/20\% \\
\midrule
carpet      & 0.770/0.667/0.582 & 0.971/\underline{0.970}/\underline{0.969} & 0.805/0.652/0.651 & 0.900/0.851/0.845 & 0.689/0.687/0.685 & \underline{0.972}/0.968/0.960 & \textbf{0.981}/\textbf{0.975}/\textbf{0.974} \\

bottle      & 0.773/0.680/0.529 & 0.969/0.965/0.962 & \textbf{0.989}/\textbf{0.984}/\textbf{0.981} & \underline{0.975}/\underline{0.972}/\underline{0.966} & 0.883/0.882/0.881 & 0.890/0.880/0.880 & 0.927/0.926/0.914 \\

cable       & 0.756/0.750/0.660 & 0.847/0.788/0.787 & 0.653/0.645/0.644 & 0.859/0.773/0.758 & 0.835/0.835/0.834 & \underline{0.893}/\underline{0.892}/\textbf{0.890} & \textbf{0.916}/\textbf{0.899}/\underline{0.869} \\

capsule     & 0.823/0.652/0.537 & 0.983/0.979/0.975 & \textbf{0.983}/\textbf{0.979}/\textbf{0.964} & 0.879/0.812/0.810 & 0.766/0.766/0.766 & \underline{0.917}/\underline{0.898}/\underline{0.894} & 0.883/0.879/0.869 \\

zipper      & 0.847/0.710/0.703 & \underline{0.977}/\underline{0.974}/\underline{0.973} & \textbf{0.987}/\textbf{0.986}/\textbf{0.985} & 0.966/0.936/0.932 & 0.555/0.551/0.550 & 0.904/0.900/0.879 & 0.969/0.965/0.962 \\

wood        & 0.760/0.679/0.448 & 0.888/0.868/0.859 & \textbf{0.960}/\textbf{0.956}/\textbf{0.931} & 0.905/0.887/\underline{0.885} & 0.720/0.718/0.717 & 0.831/0.811/0.810 & \underline{0.917}/\underline{0.889}/0.861 \\

transistor  & 0.798/0.783/0.633 & \underline{0.913}/0.874/0.822 & 0.878/\underline{0.876}/0.766 & 0.613/0.585/0.584 & 0.842/0.836/\underline{0.835} & 0.863/0.854/0.817 & \textbf{0.922}/\textbf{0.921}/\textbf{0.900} \\

toothbrush  & 0.835/0.826/0.642 & 0.934/0.929/0.900 & \textbf{0.981}/\underline{0.971}/\underline{0.968} & \underline{0.979}/\textbf{0.972}/\textbf{0.971} & 0.722/0.722/0.717 & 0.898/0.897/0.884 & 0.926/0.918/0.912 \\

tile        & 0.844/0.697/0.572 & 0.852/0.831/0.831 & \textbf{0.943}/\textbf{0.935}/0.707 & \underline{0.929}/0.901/\underline{0.872} & 0.727/0.726/0.723 & 0.858/0.808/0.800 & 0.923/\underline{0.920}/\textbf{0.904} \\

screw       & 0.799/0.709/0.564 & 0.966/\underline{0.965}/\underline{0.964} & \textbf{0.983}/\textbf{0.976}/\textbf{0.970} & \underline{0.981}/0.914/0.912 & 0.648/0.647/0.645 & 0.876/0.860/0.858 & 0.828/0.807/0.792 \\

pill        & 0.780/0.774/0.554 & 0.924/0.903/0.901 & \underline{0.934}/\textbf{0.930}/0.892 & \textbf{0.941}/\underline{0.927}/0.855 & 0.863/0.863/0.861 & 0.895/0.892/0.887 & 0.921/0.916/\underline{0.912} \\

metal\_nut  & 0.846/0.743/0.601 & 0.859/0.853/0.851 & \textbf{0.951}/\underline{0.918}/\underline{0.915} & 0.925/0.913/0.825 & 0.899/0.892/0.889 & 0.874/0.873/0.854 & \underline{0.936}/\textbf{0.923}/\textbf{0.920} \\

leather     & 0.741/0.710/0.578 & 0.983/0.979/0.977 & \textbf{0.999}/\textbf{0.999}/\textbf{0.994} & \underline{0.995}/\underline{0.992}/\underline{0.992} & 0.656/0.654/0.652 & 0.985/0.984/0.982 & 0.991/0.991/0.990 \\

hazelnut    & 0.756/0.476/0.338 & 0.957/0.953/0.889 & \textbf{0.981}/\textbf{0.981}/\textbf{0.969} & \underline{0.971}/0.907/0.901 & 0.924/0.924/0.921 & 0.939/0.934/\underline{0.930} & 0.962/\underline{0.954}/0.915 \\

grid        & 0.725/0.545/0.414 & 0.832/0.824/0.720 & \textbf{0.982}/\textbf{0.985}/\textbf{0.981} & \underline{0.978}/\underline{0.954}/\underline{0.939} & 0.608/0.602/0.601 & 0.899/0.898/0.892 & 0.938/0.934/0.934 \\
\textbf{Average} &
0.790/0.693/0.557 &
0.924/0.910/\underline{0.892} &
\textbf{0.934}/0.918/0.888 &
0.920/0.886/0.870 &
0.756/0.754/0.752 &
0.899/0.889/0.881 &
\underline{0.929}/\textbf{0.921}/\textbf{0.909} \\
\midrule
VisA Classes & 10\%/15\%/20\% & 10\%/15\%/20\% & 10\%/15\%/20\% & 10\%/15\%/20\% & 10\%/15\%/20\% & 10\%/15\%/20\% & 10\%/15\%/20\% \\
\midrule

capsules &
0.594/0.313/0.200 &
0.851/\underline{0.849}/0.749 &
\textbf{0.961}/\textbf{0.945}/\textbf{0.943} &
\underline{0.880}/0.847/\underline{0.765} &
0.547/0.542/0.538 &
0.685/0.665/0.611 &
0.759/0.742/0.738 \\

candle &
0.512/0.417/0.306 &
\textbf{0.987}/\textbf{0.985}/\textbf{0.984} &
\underline{0.983}/\underline{0.976}/\underline{0.971} &
0.910/0.859/0.763 &
0.630/0.624/0.610 &
0.746/0.741/0.736 &
0.828/0.820/0.818 \\

cashew &
0.813/0.531/0.273 &
\textbf{0.984}/\textbf{0.975}/\textbf{0.967} &
0.847/0.829/0.813 &
0.685/0.676/0.663 &
\underline{0.943}/\underline{0.942}/\underline{0.941} &
0.710/0.695/0.683 &
0.750/0.740/0.738 \\

chewinggum &
0.653/0.615/0.340 &
\underline{0.967}/\underline{0.964}/\underline{0.931} &
\textbf{0.972}/\textbf{0.968}/\textbf{0.960} &
0.942/0.923/0.873 &
0.778/0.767/0.758 &
0.916/0.905/0.901 &
0.922/0.911/0.905 \\

macaroni1 &
0.555/0.478/0.470 &
\underline{0.986}/\underline{0.980}/0.974 &
\textbf{0.989}/\textbf{0.985}/\textbf{0.980} &
0.985/0.979/\underline{0.975} &
0.540/0.531/0.511 &
0.823/0.815/0.811 &
0.889/0.887/0.859 \\

macaroni2 &
0.435/0.345/0.320 &
0.933/0.883/0.857 &
\textbf{0.982}/\textbf{0.980}/\textbf{0.973} &
\underline{0.979}/\underline{0.975}/\underline{0.969} &
0.476/0.466/0.451 &
0.804/0.783/0.737 &
0.835/0.821/0.811 \\

fryum &
0.830/0.724/0.533 &
\textbf{0.933}/\textbf{0.926}/\underline{0.902} &
0.868/0.865/0.845 &
0.756/0.699/0.662 &
\underline{0.916}/\underline{0.915}/\textbf{0.905} &
0.855/0.835/0.827 &
0.891/0.881/0.872 \\

pipe\_fryum &
0.780/0.558/0.393 &
\textbf{0.990}/\textbf{0.989}/\textbf{0.984} &
0.948/0.926/0.911 &
0.764/0.743/0.526 &
\underline{0.960}/\underline{0.960}/\underline{0.958} &
0.812/0.789/0.778 &
0.911/0.909/0.905 \\

pcb1 &
0.569/0.468/0.324 &
\textbf{0.986}/\textbf{0.974}/\textbf{0.968} &
\underline{0.922}/\underline{0.911}/\underline{0.898} &
0.512/0.487/0.459 &
0.743/0.741/0.739 &
0.733/0.633/0.605 &
0.862/0.836/0.820 \\

pcb2 &
0.502/0.449/0.304 &
\textbf{0.981}/\textbf{0.973}/\textbf{0.968} &
\underline{0.960}/\underline{0.948}/\underline{0.933} &
0.767/0.743/0.679 &
0.579/0.571/0.569 &
0.683/0.675/0.631 &
0.876/0.859/0.855 \\

pcb3 &
0.482/0.321/0.296 &
\underline{0.982}/\underline{0.973}/\textbf{0.966} &
\textbf{0.985}/\textbf{0.982}/\underline{0.938} &
0.797/0.796/0.790 &
0.644/0.641/0.637 &
0.651/0.586/0.546 &
0.868/0.858/0.845 \\

pcb4 &
0.531/0.493/0.474 &
\underline{0.944}/\underline{0.925}/\underline{0.916} &
\textbf{0.945}/\textbf{0.943}/\textbf{0.939} &
0.738/0.716/0.772 &
0.740/0.732/0.728 &
0.852/0.849/0.817 &
0.907/0.903/0.890 \\

\textbf{Average} &
0.605/0.476/0.353 &
\textbf{0.960}/\textbf{0.950}/\textbf{0.931} &
\underline{0.947}/\underline{0.938}/\underline{0.925} &
0.809/0.787/0.741 &
0.708/0.703/0.696 &
0.777/0.748/0.724 &
0.858/0.847/0.838 \\
\bottomrule
\end{tabular}
}
\end{table*}


\subsubsection{Anomaly Detection} We report image-level AUROC results on the MVTec AD and VisA datasets under different contamination ratios in Table~\ref{tab:image_roc}. Across all MVTec categories, the proposed method consistently achieves the highest average performance, with particularly pronounced gains at higher contamination levels (15\% and 20\%). This highlights the effectiveness of the proposed learning strategy in maintaining reliable detection performance when the training data is heavily corrupted.

Compared to segmentation-based methods such as DestSeg, which exhibit strong performance on texture categories (e.g., grid, leather), our approach demonstrates superior robustness across both texture and object-centric classes. In particular, while PatchCore, DestSeg and PaDiM often deteriorate significantly as contamination increases, the proposed method maintains stable and high AUROC scores, underscoring its resilience to contamination.

On the more challenging VisA dataset, all methods experience a noticeable drop in absolute performance. However, the proposed approach consistently outperforms competing baselines across all contamination levels, achieving the best average AUROC. This shows that our method generalizes well across diverse and complex anomaly patterns.

\subsubsection{Anomaly Localization} We summarize pixel-level localization results in Table~\ref{tab:pixel_roc}. On MVTec, segmentation-based methods such as DestSeg often achieve the highest pixel-level AUROC, which is expected given their explicit optimization for dense pixel-wise supervision. Despite not being explicitly optimized for dense pixel-wise segmentation, the proposed method remains competitive, frequently ranking among the top-performing approaches and exhibiting strong robustness as contamination increases.

For the VisA dataset, localization is substantially more challenging for all methods. The anomalies in VisA are often irregular, sparse, and exhibit weak or ambiguous boundaries. As a result, methods explicitly optimized for segmentation retain an advantage in terms of pixel-level AUROC. In contrast, our approach focuses on learning deviation-consistent anomaly scores derived from explainability signals, rather than optimizing pixel-wise boundaries. Consequently, while pixel-level AUROC is lower, the generated localization maps still provide meaningful spatial cues that support anomaly interpretation.


\begin{figure*}[ht]
    \centering
    \begin{subfigure}[b]{0.48\linewidth}
        \centering
        \includegraphics[width=\linewidth]{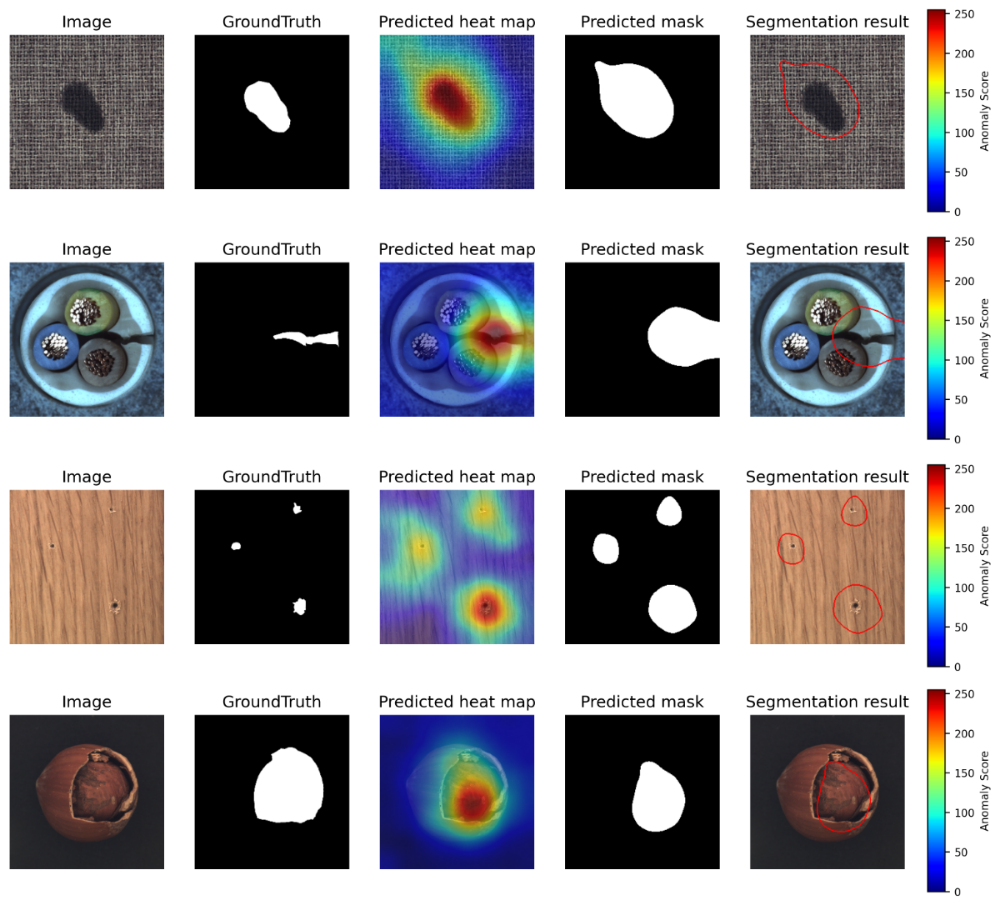}
        \caption{Example localization outputs on the MVTec AD dataset.}
        \label{fig:mvtec_qualitative}
    \end{subfigure}
     \hfill
    \begin{subfigure}[b]{0.48\linewidth}
        \centering
        \includegraphics[width=\linewidth]{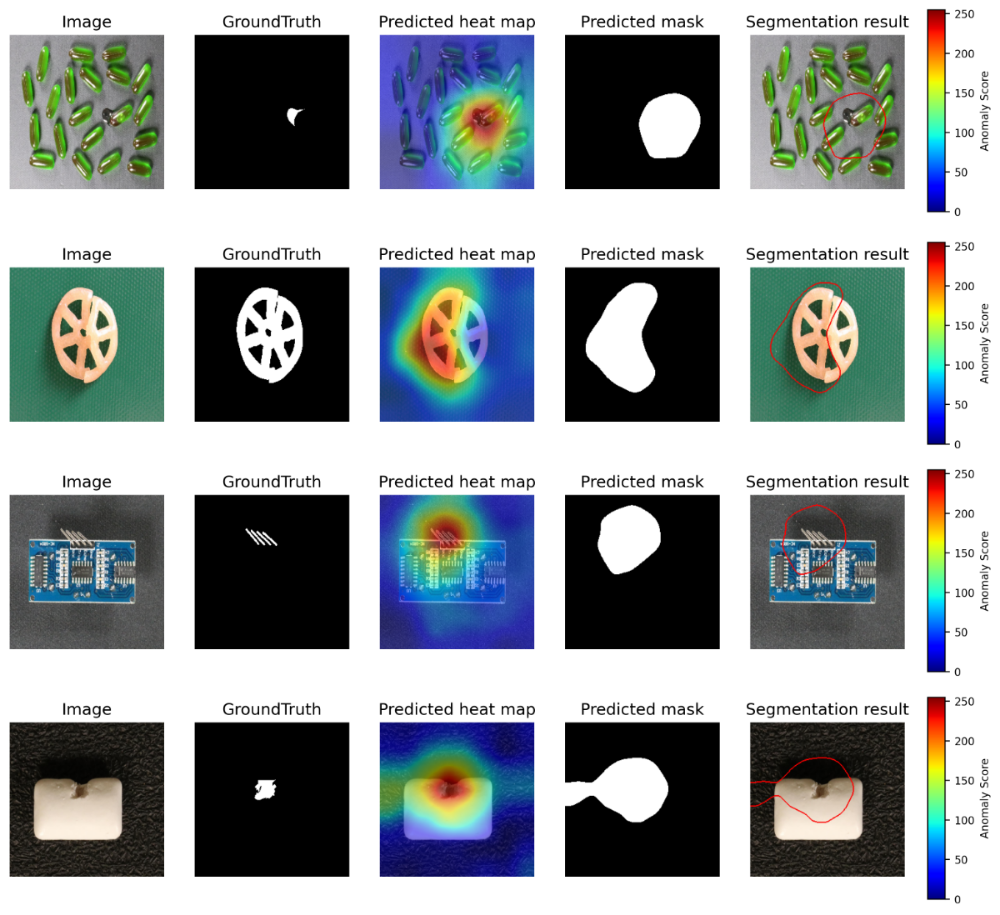}
        \caption{Example localization outputs on the VisA dataset.}
        \label{fig:visa_qualitative}
    \end{subfigure}

    \caption{Anomaly localization examples showing input images, ground-truth masks, predicted heatmaps, predicted masks, and segmentation outputs for both MVTec AD and VisA.}
    \label{fig:qualitative_results}
\end{figure*}

\subsubsection{Qualitative Results}

 We present qualitative anomaly localization results on representative samples from the MVTec AD and VisA datasets in Figure~\ref{fig:qualitative_results}. For each example, we visualize the input image, ground-truth anomaly mask, predicted anomaly heatmap, corresponding binary localization output, and segmentation outputs.

As shown in Fig.~\ref{fig:mvtec_qualitative}, the proposed method produces spatially coherent anomaly responses that closely align with ground-truth defect regions across diverse object and texture categories. The predicted heatmaps emphasize salient anomalous areas while suppressing normal background responses, yielding clean and well-localized segmentation outputs across both compact and irregular defect patterns.

Fig.~\ref{fig:visa_qualitative} illustrates localization examples on the VisA dataset, which is known to contain more complex, diverse, and less visually homogeneous anomalies. Compared to MVTec, anomaly regions in VisA are often fragmented or diffused. In this setting, the proposed method still highlights anomalous regions with meaningful spatial focus, although the resulting heatmaps are generally less compact. This observation is consistent with the quantitative localization results, where pixel-level AUROC on VisA is lower.

Overall, these qualitative results demonstrate that the proposed approach produces interpretable and stable localization maps without explicitly optimizing a pixel-level segmentation objective, complementing its strong image-level detection performance.

\begin{figure*}[ht]
  \centering
    \includegraphics[width=0.9
    \linewidth]{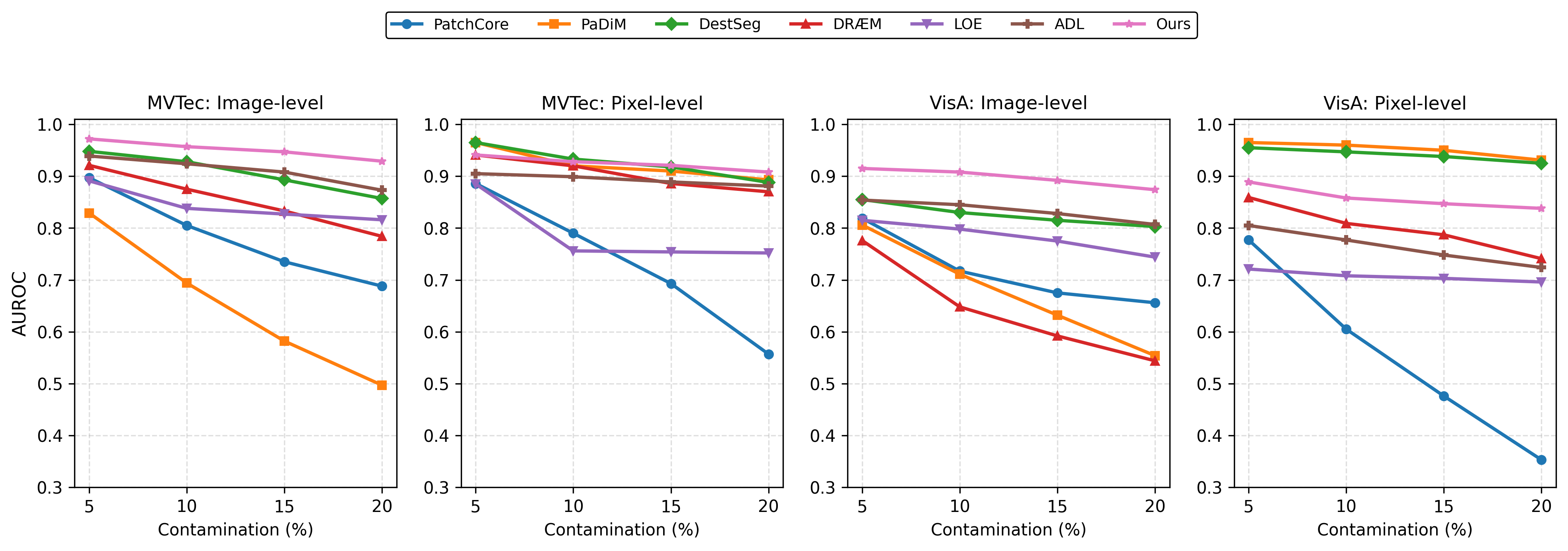}
    \caption{Robustness evaluation: Image-level and pixel-level AUROC under varying anomaly contamination rates on the MVTec and VisA datasets.}
    \label{fig:robust_roc}
  
\end{figure*}

\subsubsection{Robustness to Contamination}
We evaluate robustness under increasing anomaly contamination, reporting image-level and pixel-level AUROC on MVTec AD and VisA. As depicted in Figure~\ref{fig:robust_roc}, when contamination increases from 5\% to 20\%, most baselines exhibit a noticeable performance drop, particularly feature-based methods such as PatchCore and PaDiM, which are sensitive to shifts in the normal feature distribution.
On MVTec, the proposed method consistently achieves the highest image-level performance across all contamination ratios and shows only a mild degradation as contamination increases. Similar behavior is observed for pixel-level localization, where the proposed approach maintains stable performance while several baselines degrade substantially, especially at higher contamination levels.
VisA presents a more challenging setting due to increased visual diversity and less well-defined anomaly boundaries. While all methods experience performance degradation under increasing contamination, the proposed method remains consistently competitive and exhibits stable behavior at both the image-level and the pixel-level. The performance gap becomes more pronounced at higher contamination ratios, indicating improved robustness under severe noise.
Overall, the results demonstrate that the proposed method degrades gracefully under increasing contamination, providing stable detection and localization performance.

\begin{figure*}[ht]
    \centering
    \begin{subfigure}[b]{0.24\linewidth}
        \centering
        \includegraphics[width=\linewidth]{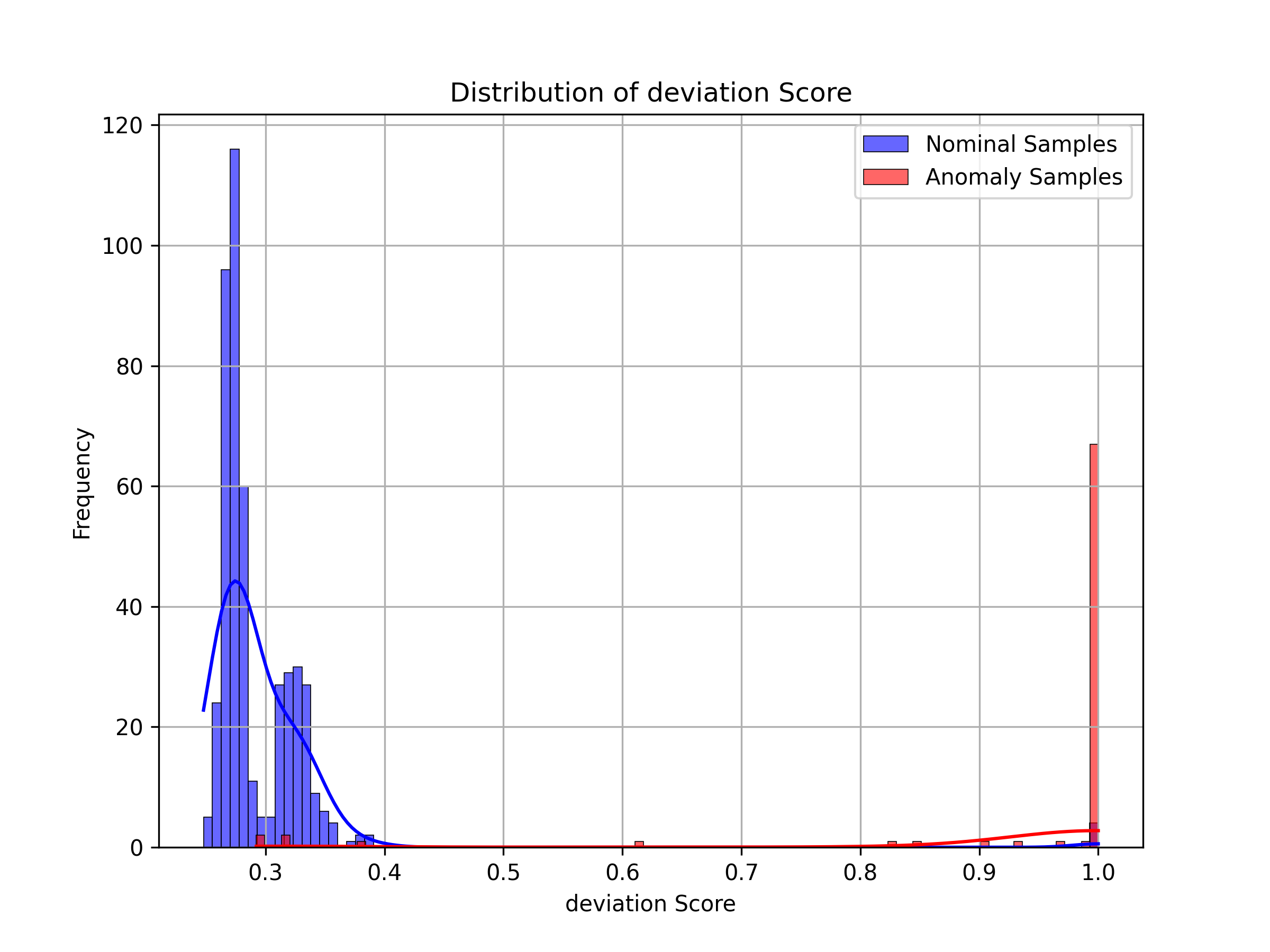}
        \caption{Deviation Score}
        \label{fig:deviation_score_dist}
    \end{subfigure}
    \hfill
    \begin{subfigure}[b]{0.24\linewidth}
        \centering
        \includegraphics[width=\linewidth]{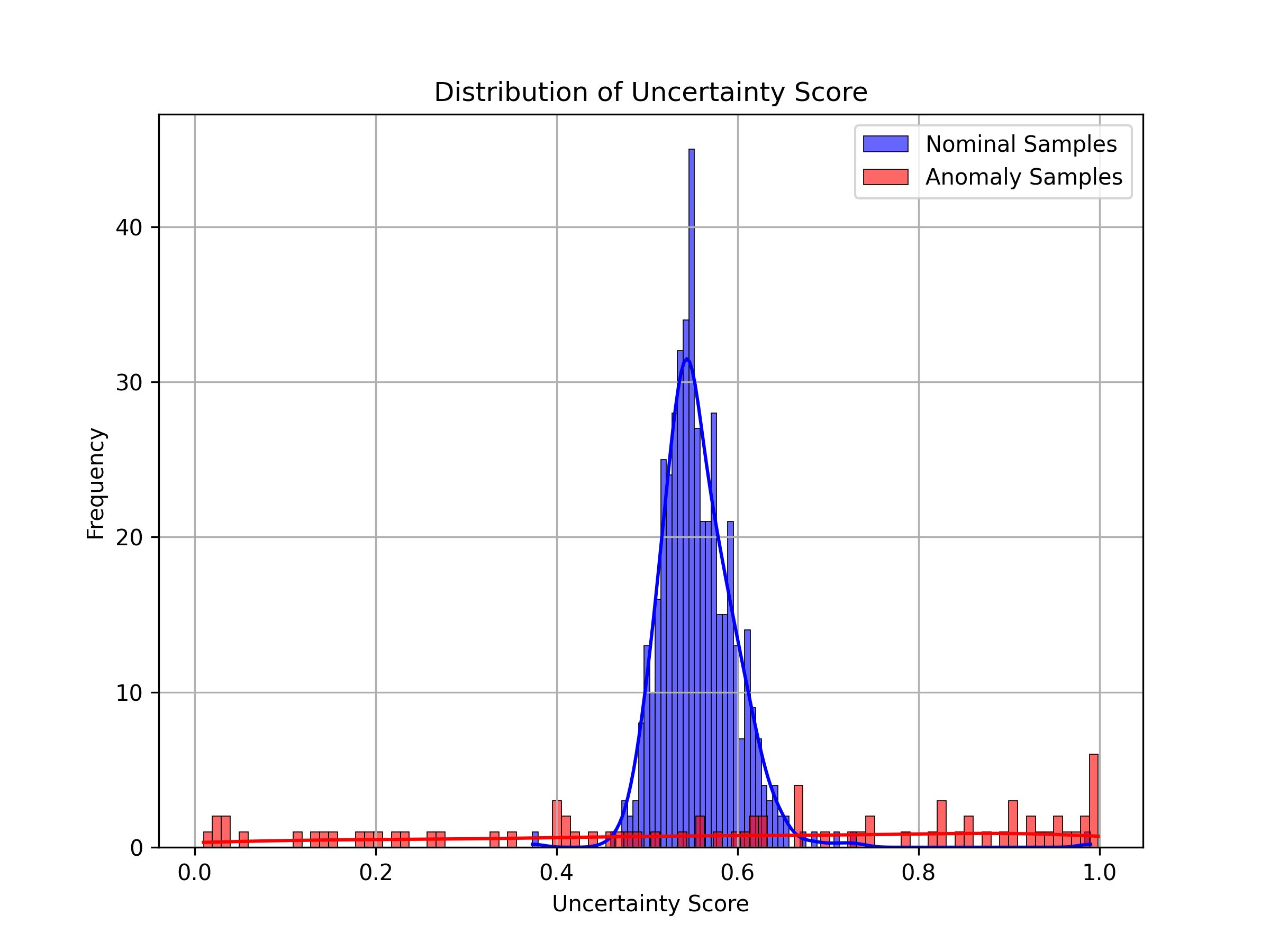}
        \caption{Uncertainty Score}
        \label{fig:entropy_score_dist}
    \end{subfigure}
    \hfill
    \begin{subfigure}[b]{0.24\linewidth}
        \centering
        \includegraphics[width=\linewidth]{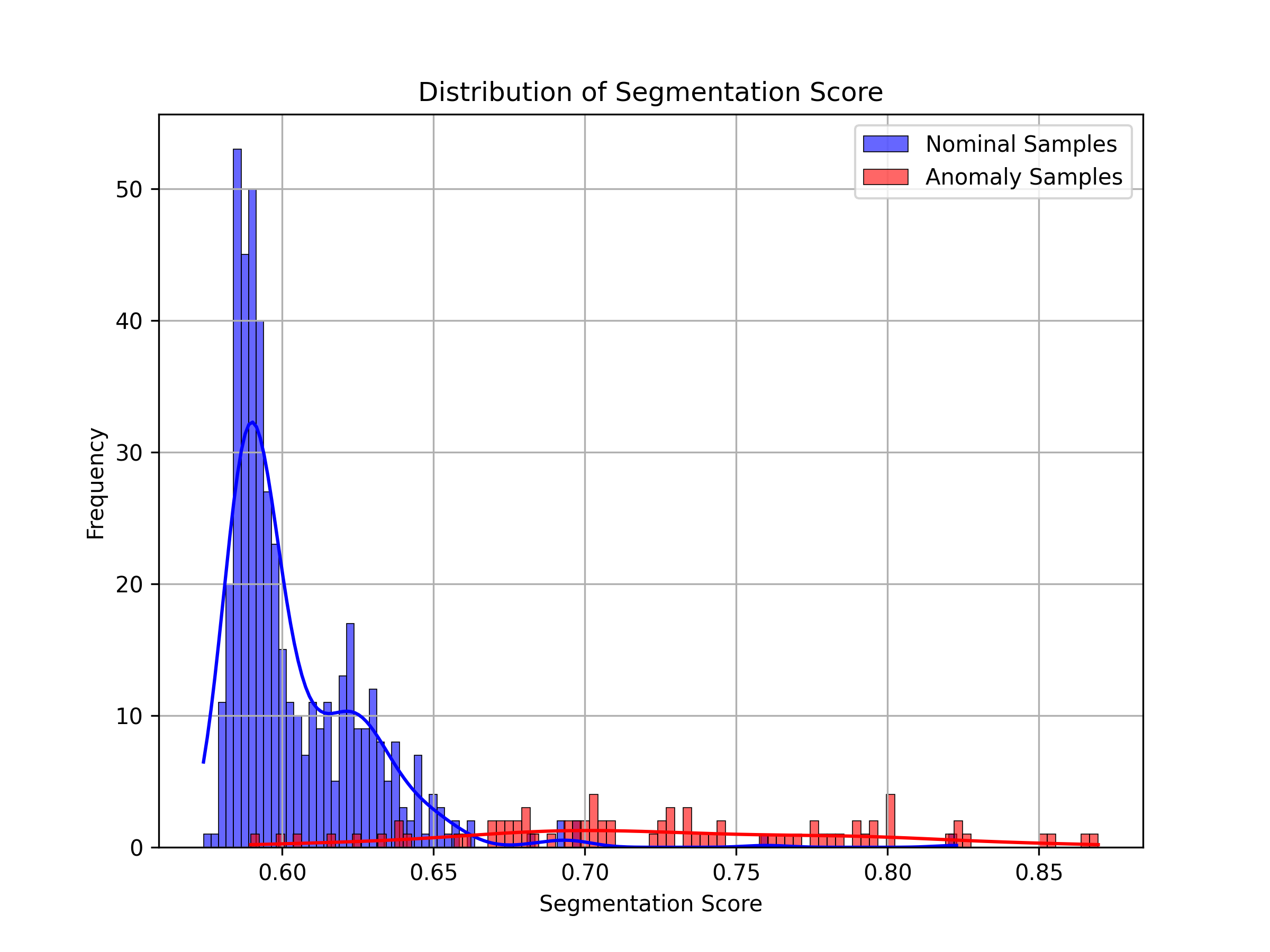}
        \caption{Segmentation Score}
        \label{fig:seg_score_dist}
    \end{subfigure}
    \hfill
    \begin{subfigure}[b]{0.24\linewidth}
        \centering
        \includegraphics[width=\linewidth]{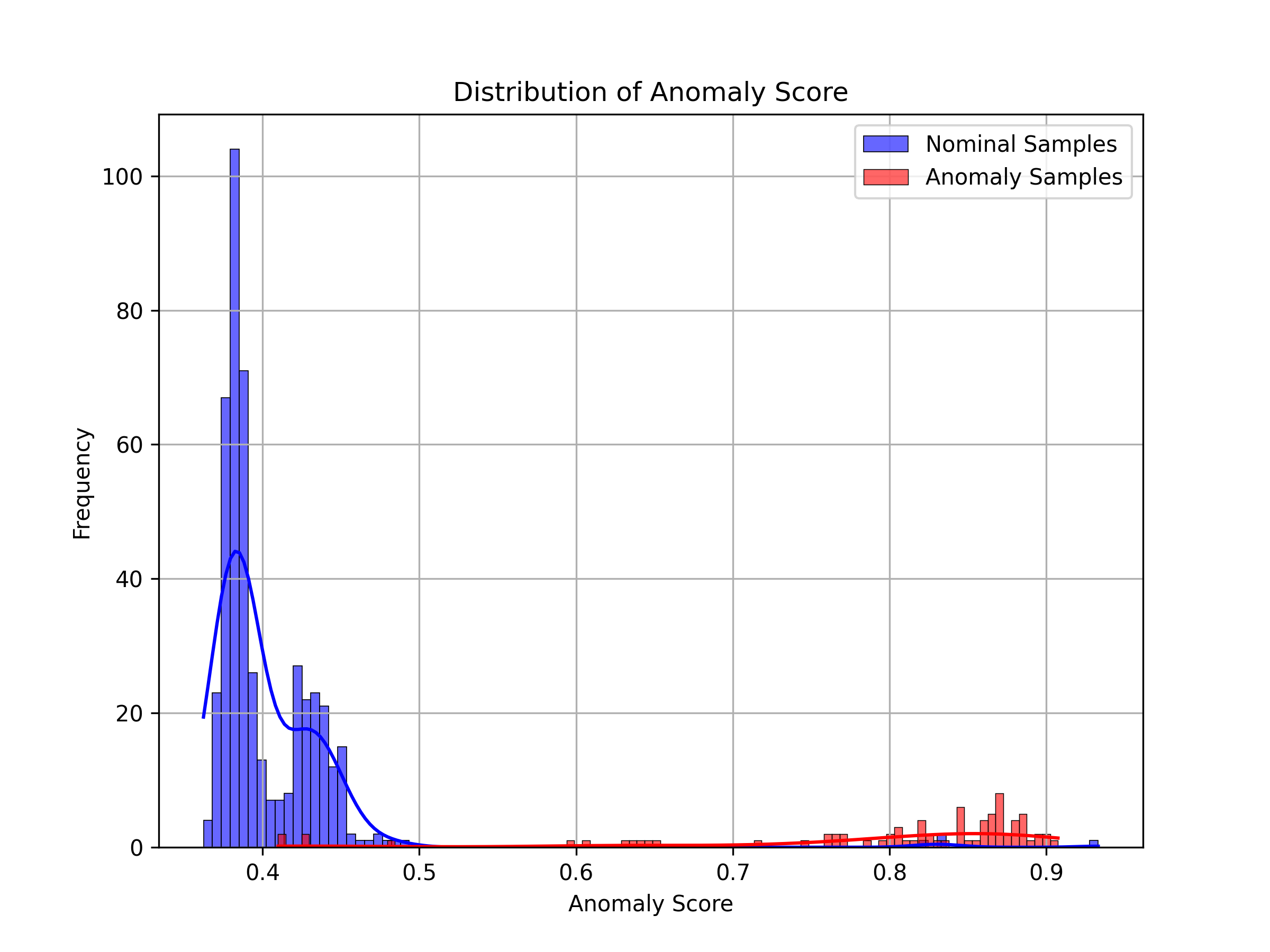}
        \caption{Anomaly Score (Combined)}
        \label{fig:anomaly_score_dist}
    \end{subfigure}

    \caption{Distributions of deviation, uncertainty, segmentation, and fused anomaly scores for nominal and anomalous samples. Each score highlights different aspects of anomalous behavior, illustrating their separability.}
    \label{fig:score_distributions}
\end{figure*}

\subsubsection{Multi-Signal Perspectives on Anomalous Behavior}

In Figure~\ref{fig:score_distributions}, we show the distributions of the individual anomaly cues: deviation, uncertainty, and segmentation scores; together with their fused anomaly score, for nominal and anomalous samples. Each score captures a complementary aspect of abnormal behavior and exhibits different separation characteristics.
The deviation score shows a strong concentration for nominal samples with a long tail for anomalous ones. The uncertainty score displays moderate overlap between nominal and anomalous samples, indicating that uncertainty score alone is not sufficient for reliable discrimination, specifically under contamination. In contrast, the segmentation score yields improved separation by emphasizing spatial consistency of anomalous regions, although some overlap remains due to ambiguous or diffused anomalies.
Importantly, the combined anomaly score effectively integrates these heterogeneous cues, resulting in a clearer margin between normal and anomalous samples. The fused score suppresses spurious responses present in individual signals while enhancing consistent anomaly evidence, leading to a more discriminative and stable scoring function. 

\subsubsection{Sensitivity to Contamination}
\begin{figure*}[ht]
    \centering
    \begin{subfigure}[b]{0.4\linewidth}
        \centering
        \includegraphics[width=\linewidth]{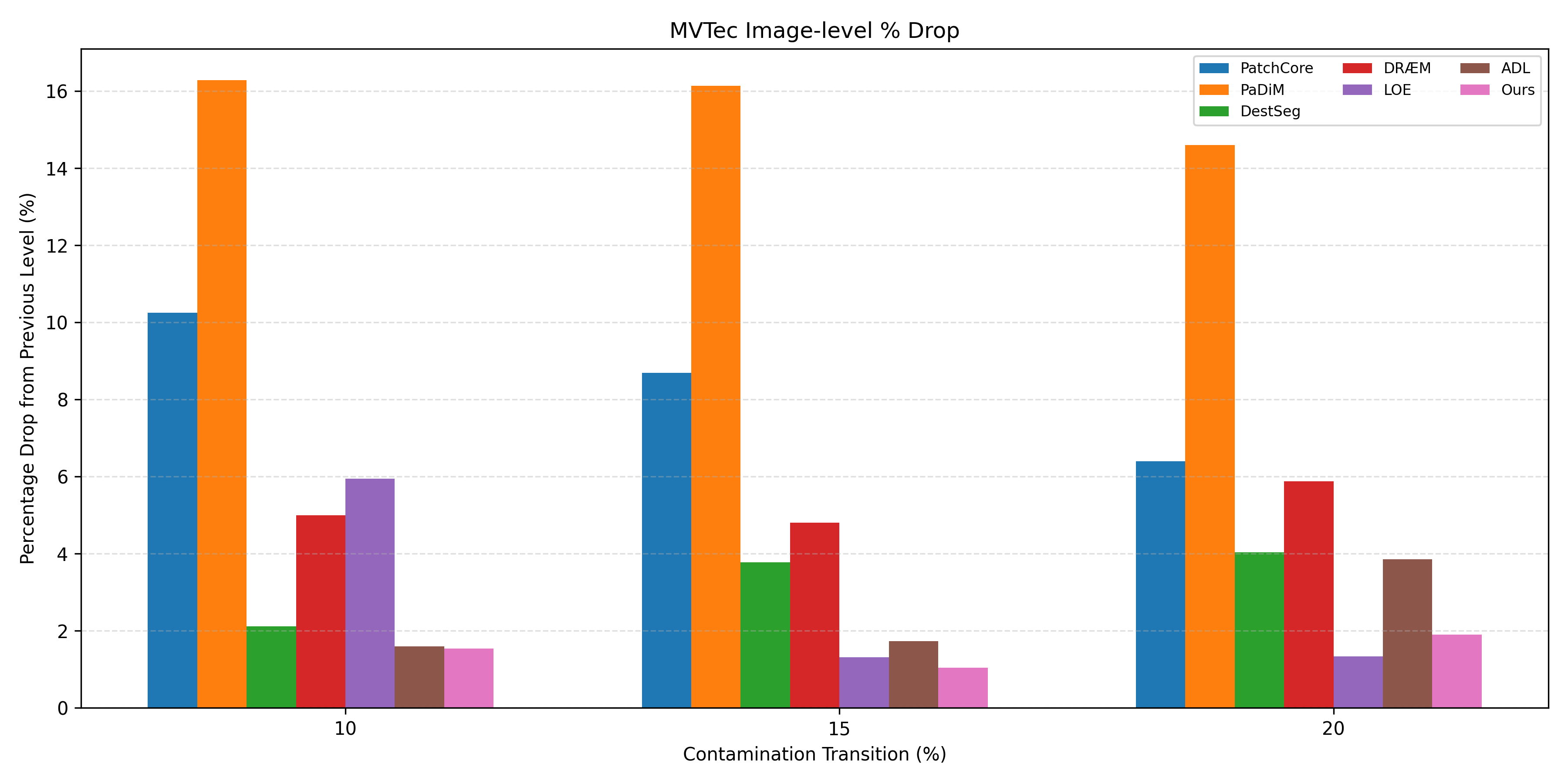}
        \caption{MVTec: Image-level AUROC drop}
        \label{fig:mvtec_image_drop}
    \end{subfigure}
    \hfill
    \begin{subfigure}[b]{0.4\linewidth}
        \centering
        \includegraphics[width=\linewidth]{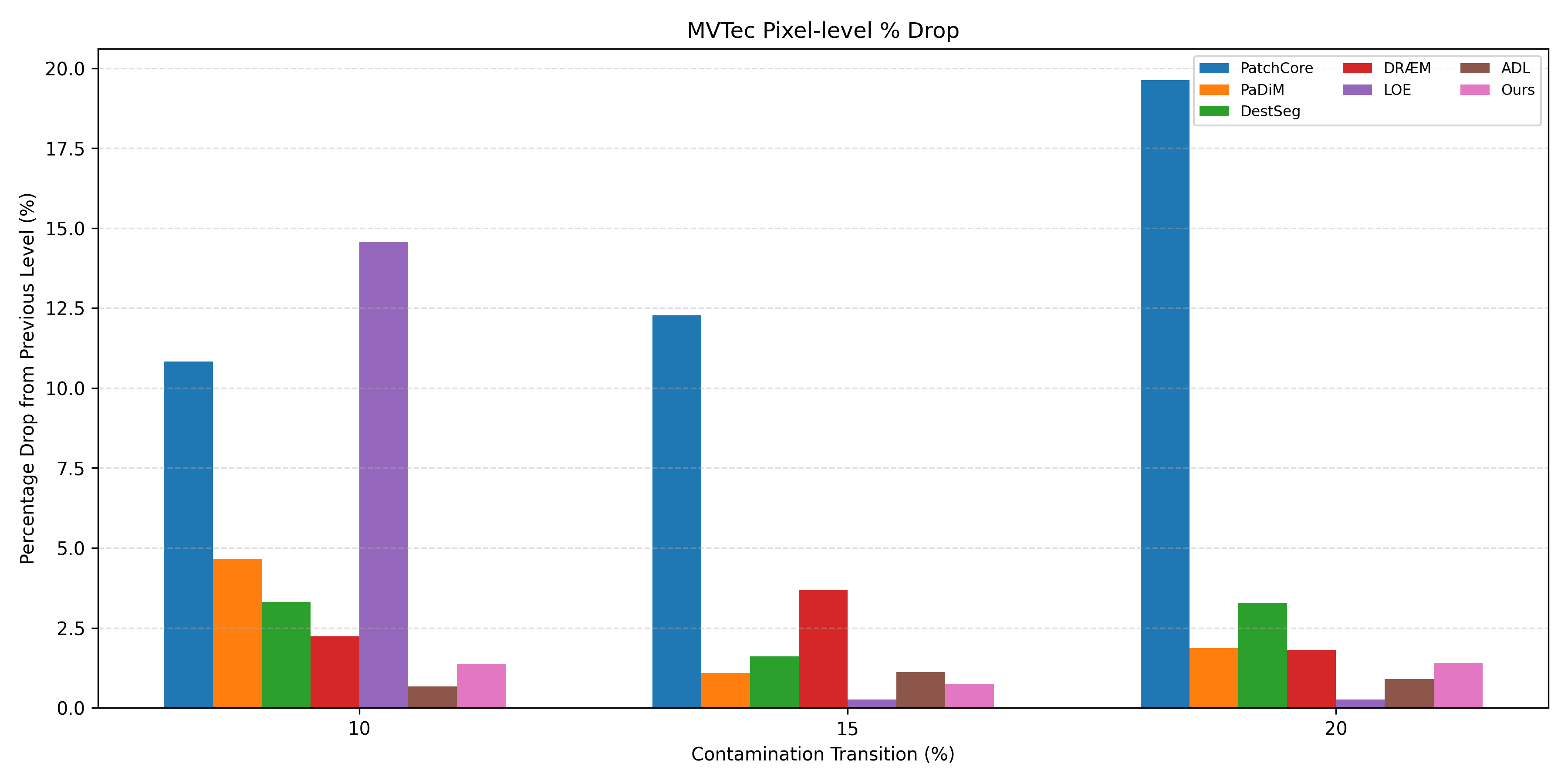}
        \caption{MVTec: Pixel-level AUROC drop}
        \label{fig:mvtec_pixel_drop}
    \end{subfigure}


    \begin{subfigure}[b]{0.4\linewidth}
        \centering
        \includegraphics[width=\linewidth]{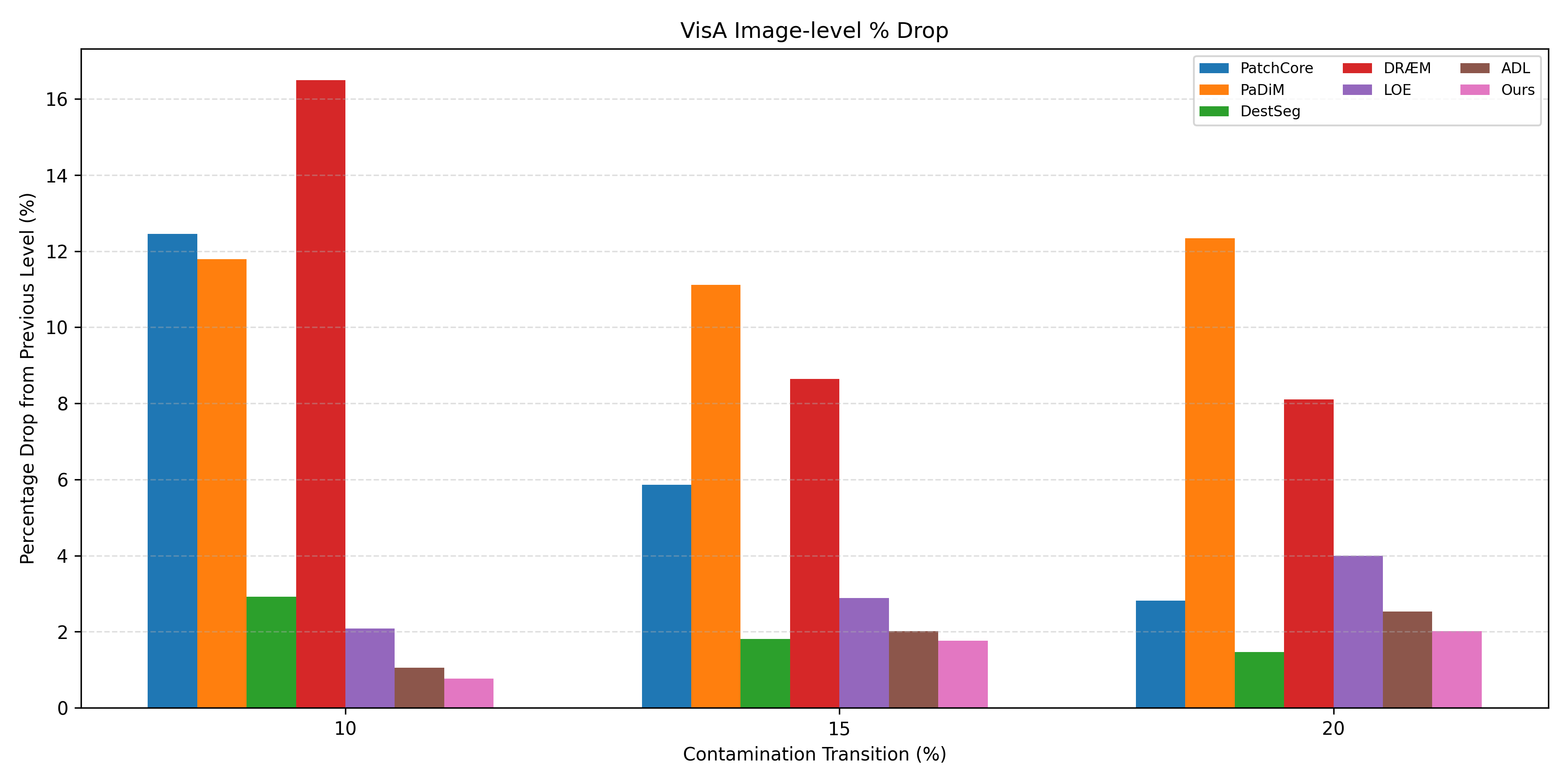}
        \caption{VisA: Image-level AUROC drop}
        \label{fig:visa_image_drop}
    \end{subfigure}
    \hfill
    \begin{subfigure}[b]{0.4\linewidth}
        \centering
        \includegraphics[width=\linewidth]{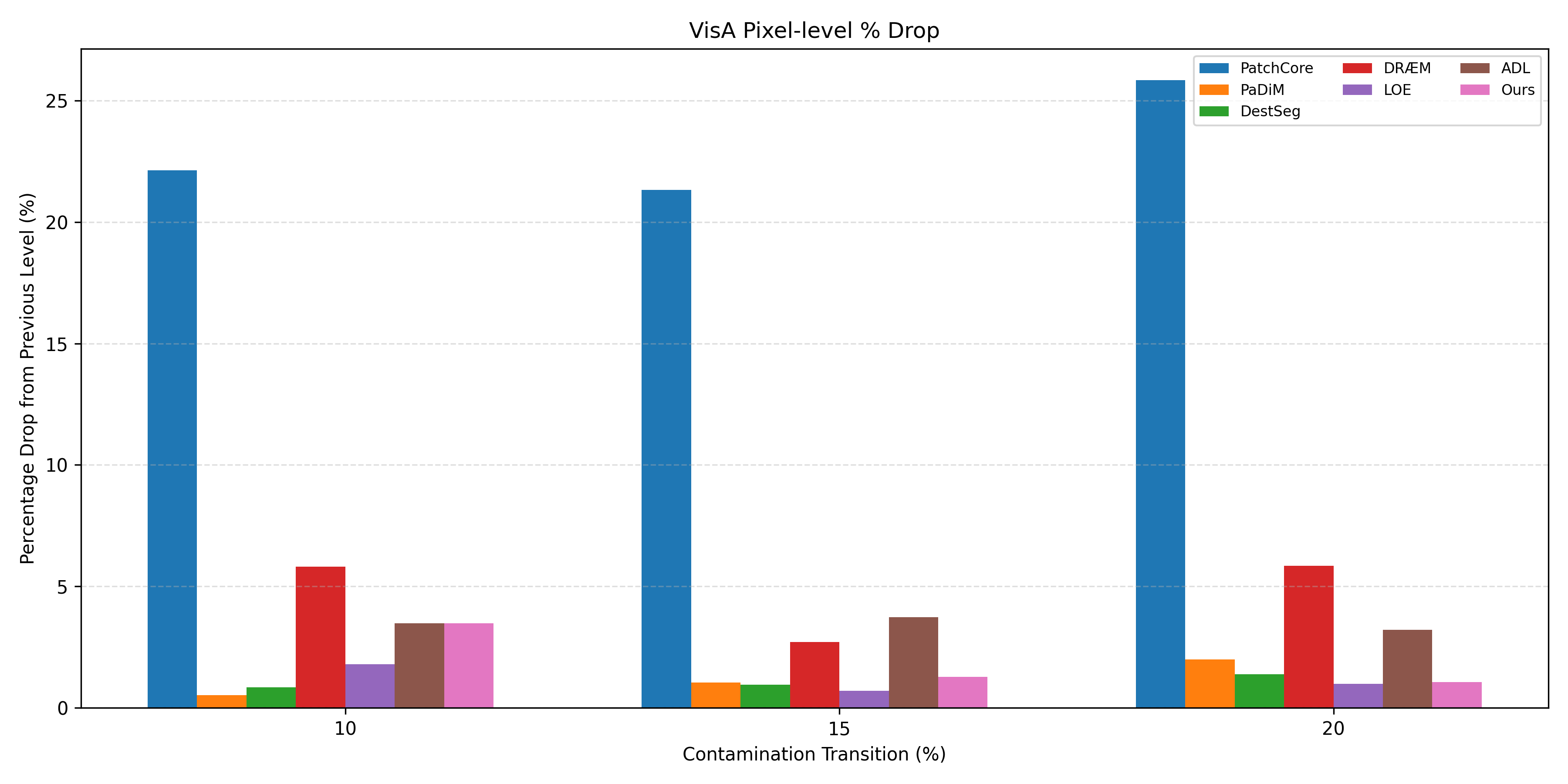}
        \caption{VisA: Pixel-level AUROC drop}
        \label{fig:visa_pixel_drop}
    \end{subfigure}
    
    \caption{Drop (\%) in image and pixel-level AUROC between consecutive contamination levels (5\%$\rightarrow$10\%, 10\%$\rightarrow$15\%, and 15\%$\rightarrow$20\%). The plots summarize sensitivity to contamination on (a)–(b) MVTec and (c)–(d) VisA.}
    \label{fig:contamination_drop_overview}
\end{figure*}

To assess robustness under increasing contamination, we measure the relative AUROC drop between consecutive contamination levels (5\%$\rightarrow$10\%, 10\%$\rightarrow$15\%, and 15\%$\rightarrow$20\%) at both image and pixel-levels on MVTec and VisA. The results are illustrated in Figure~\ref{fig:contamination_drop_overview}. Lower performance drop indicates lower sensitivity to increasing contamination.

At the pixel-level, PatchCore exhibits the most severe degradation, with AUROC drops exceeding 20\% on VisA and approaching 20\% on MVTec, indicating strong sensitivity to contaminated training data. 
At the image-level, PaDiM shows the largest performance degradation across contamination transitions, while DR{\AE}M exhibits moderate but less consistent declines, particularly on VisA. Notably, LOE and ADL exhibit reasonable robustness both at image-level and pixel-level.
In contrast, the proposed method demonstrates consistently small performance drops across all transitions and evaluation levels. 
On MVTec, both image and pixel-level degradation remains around or below 2\% as contamination increases. 
On the more challenging VisA dataset, the proposed method maintains stable behavior, with drops generally limited to approximately 1--2\%, substantially lower than competing approaches.
Overall, these results indicate that the proposed multi-cue anomaly scoring framework not only achieves strong performance but also preserves stability under increasing contamination, particularly for fine-grained localization.

\subsubsection{Sample Efficiency}

\begin{figure*}[!ht]
    \centering
    \begin{subfigure}[b]{0.4\linewidth}
        \includegraphics[width=\linewidth]{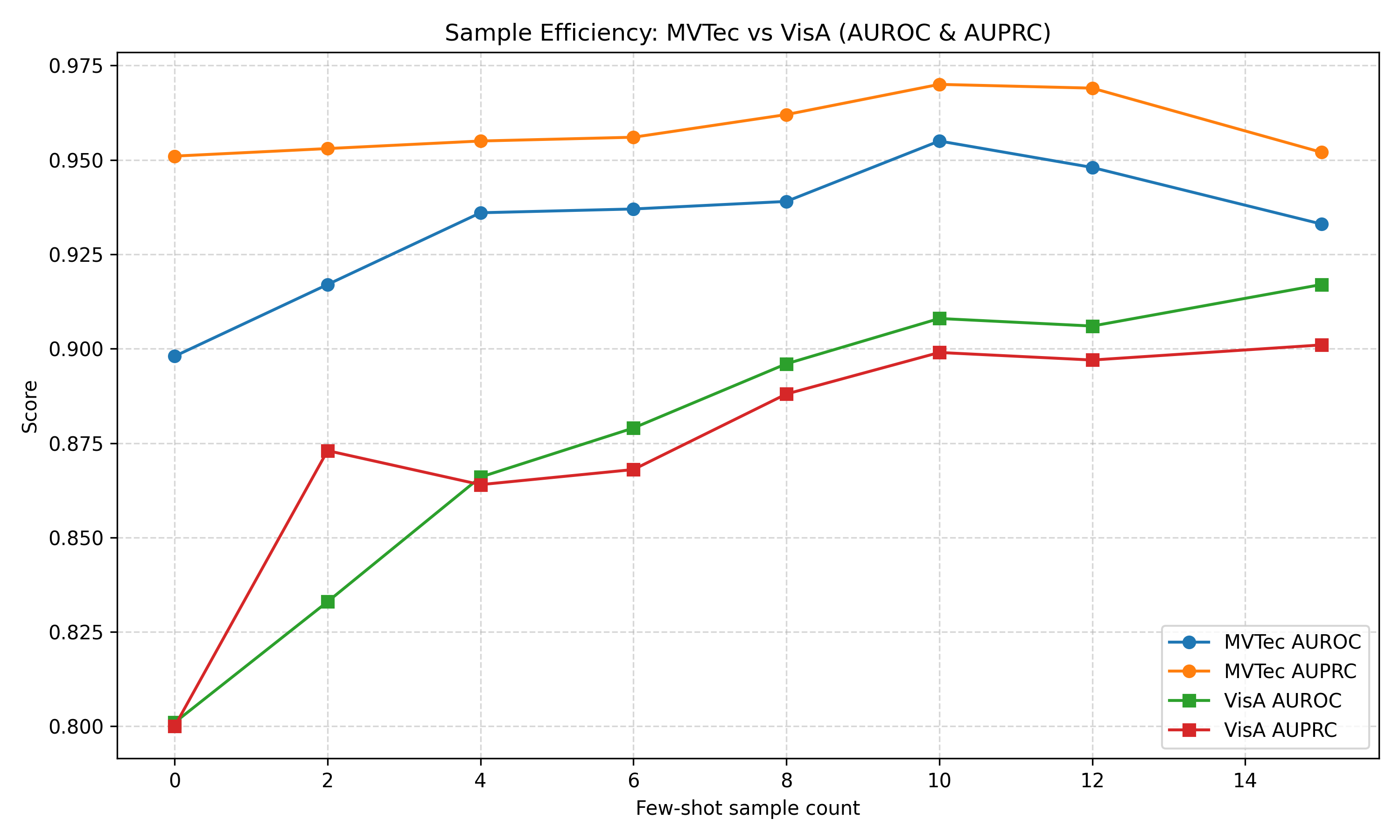}
        \caption{Sample-Efficiency Curve}
        \label{fig:sample_eff_curve}
    \end{subfigure}
    \hfill
    \begin{subfigure}[b]{0.4\linewidth}
        \includegraphics[width=\linewidth]{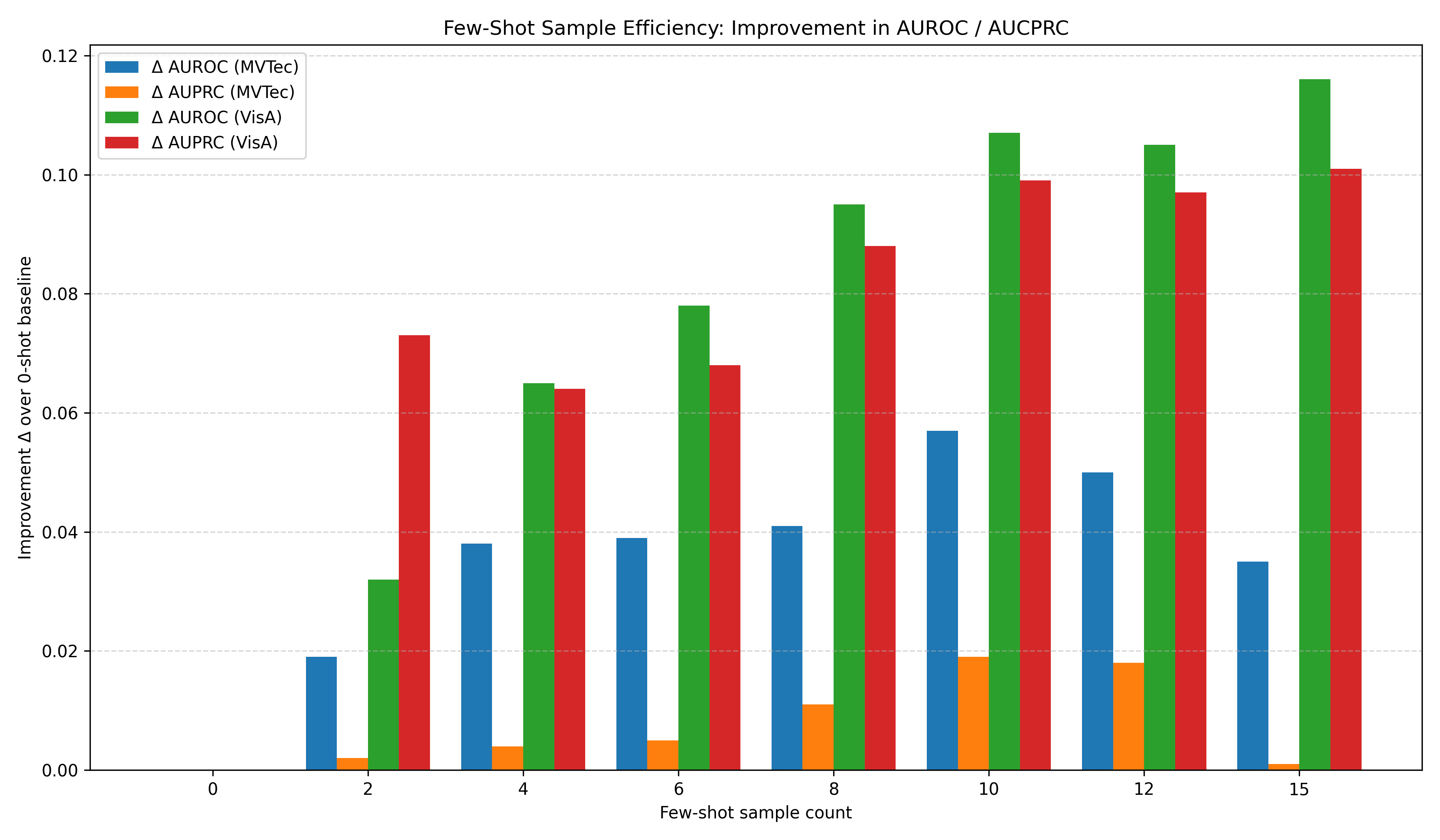}
        \caption{$\Delta$ AUROC Improvements Across Few-Shot Settings}
        \label{fig:delta_auc}
    \end{subfigure}
    
    \caption{Sample-efficiency of the proposed method.
(a) AUROC/AUPRC vs.\ number of few-shot anomaly samples.
(b) Relative AUROC gain ($\Delta$) over the 0-shot baseline.}
    \label{fig:sample_efficiency_overview}
\end{figure*}

We further assess the effect of limited anomaly supervision by varying the number of labeled anomaly samples $m$ introduced during training. Figure~\ref{fig:sample_efficiency_overview} demonstrates that the proposed method is highly sample-efficient on both the MVTec AD and VisA datasets.

As shown in Fig.~\ref{fig:sample_eff_curve}, even a very small number of labeled anomalies leads to substantial improvements over the 0-shot setting. On MVTec, performance gains saturate quickly, indicating that the model can effectively exploit minimal supervision to refine anomaly decision boundaries. On VisA dataset, performance increases more gradually but consistently.

Fig.~\ref{fig:delta_auc} additionally shows the relative improvement over the unsupervised baseline. The largest gains occur in the low-shot regime (\(m \leq 10\)), confirming that the proposed framework benefits most from limited anomaly examples. As additional samples are introduced, improvements diminish, suggesting that the model does not rely on dense anomaly supervision and avoids overfitting to a limited anomaly distribution.

In general, these results highlight that the proposed approach effectively integrates limited anomaly supervision into contamination-robust deviation learning, achieving strong performance gains with only a handful of labeled anomalies. 

\subsubsection{Ablation Study}

\begin{table}[ht]
\centering
\caption{Ablation study: Mean Image-level AUROC and AUPRC on MVTec and VisA evaluated at 10\% contamination.}
\label{tab:ablation_scores}
\scalebox{0.7}{
\begin{tabular}{lcccc}
\toprule
 & \textbf{Deviation Score} & \textbf{Uncertainty Score} & \textbf{Segmentation Score} & \textbf{Proposed} \\
\midrule
AUROC (MVTec) & 0.941 & 0.622 & 0.939 & \textbf{0.955} \\
AUPRC (MVTec)  & 0.952 & 0.808 & 0.949 & \textbf{0.970} \\
\midrule
AUROC (VisA) & 0.898 & 0.801 & 0.839 & \textbf{0.908} \\
AUPRC (VisA)  & 0.878 & 0.866 & 0.847 & \textbf{0.899} \\
\bottomrule
\end{tabular}
}
\end{table}

We analyze the contribution of individual anomaly cues and their combination within the proposed framework. Specifically, we evaluate three single-cue variants that rely exclusively on (i) deviation-based scoring, (ii) entropy-based uncertainty scoring, and (iii) segmentation-derived anomaly cues. These variants are compared against the full model, which integrates all cues through the proposed multi-signal anomaly scoring strategy.

Table~\ref{tab:ablation_scores} reports mean AUROC and AUPRC results on MVTec AD and VisA under a fixed contamination level of 10\%. When used in isolation, deviation-based scoring achieves strong performance on both datasets, confirming its effectiveness as a contamination-robust signal. Entropy-based scoring captures uncertainty patterns but exhibits substantially lower performance, particularly on MVTec, indicating limited discriminative power when used alone. Segmentation-based scoring performs competitively on MVTec but is less effective when used as a standalone anomaly signal, particularly on VisA, indicating that it is insufficient as a sole anomaly cue.

Combining all three cues yields the best performance across both datasets. The improvement is more pronounced on VisA, suggesting that leveraging multiple anomaly cues becomes increasingly important in scenarios with higher anomaly diversity and semantic complexity. These results demonstrate that the proposed tri-signal fusion formulation effectively leverages complementary information from deviation, uncertainty, and segmentation signals, leading to more reliable anomaly detection than any individual component alone.

\section{Concluding Remarks}

In this paper, we propose a contamination-robust anomaly detection framework that unifies deviation, uncertainty, and
segmentation-derived cues within a single anomaly scoring
formulation. The resulting multi-signal score enables reliable
anomaly detection and interpretable anomaly localization.
Extensive experiments on benchmarks demonstrate strong
robustness under increasing contamination, with limited performance degradation compared to existing methods. The
approach further exhibits high sample efficiency, achieving notable gains with only a few labeled anomaly examples. In summary, this work provides a practical and robust solution for
anomaly detection and localization in realistic settings where
training data may be contaminated, and supervision is limited.
Future work will explore adaptive weighting of
multiple anomaly signals and tighter integration of semantic-aware localization to further improve pixel-level precision on
complex general-purpose datasets beyond industrial inspection.





\bibliography{references}

\end{document}